\documentclass[preprint,12pt]{elsarticle}

\usepackage{amssymb}
\usepackage{amsmath}
\usepackage{subcaption}
\usepackage{dsfont}
\usepackage{float}
\usepackage{multirow}
\usepackage{bm}
\usepackage{hyperref}
\usepackage{booktabs}
\usepackage{siunitx}
\usepackage[final]{changes}

\setcitestyle{authoryear,open={(},close={)}}
\setcitestyle{authoryear} 

\journal{}

\begin{document}

\begin{frontmatter}

\title{Fault detection and diagnosis for the engine electrical system of a space launcher based on a temporal convolutional autoencoder and calibrated classifiers}

\author[onera_tls]{Luis Basora}
\author[onera_tls]{Louison Bocquet-Nouaille}
\author[onera_paris]{Elinirina Robinson}
\author[arianegroup]{Serge Le Gonidec}

\affiliation[onera_tls]{organization={ONERA DTIS, Université de         Toulouse}, addressline={F-31055 Toulouse}, city={Toulouse}, country={France}}
\affiliation[onera_paris]{organization={ONERA DTIS, Université Paris-Saclay},
addressline={Chemin de la Huniere, 91123}, city={Paris}, country={France}}
\affiliation[arianegroup]{organization={ArianeGroup SAS}, addressline={ Forêt de Vernon, 27208}, city={Vernon}, country={France}}

\begin{abstract}
In the context of the health monitoring for the next generation of reusable space launchers, we outline a first step toward developing an onboard fault detection and diagnostic capability for the electrical system that controls the engine valves. Unlike existing approaches in the literature, our solution is designed to meet a broader range of key requirements. This includes estimating confidence levels for predictions, detecting out-of-distribution (OOD) cases, and controlling false alarms. The proposed solution is based on a temporal convolutional autoencoder to automatically extract low-dimensional features from raw sensor data. Fault detection and diagnosis are respectively carried out using a binary and a multiclass classifier trained on the autoencoder latent and residual spaces. The classifiers are histogram-based gradient boosting models calibrated to output probabilities that can be interpreted as confidence levels. A relatively simple technique, based on inductive conformal anomaly detection, is used to identify OOD data. We leverage other simple yet effective techniques, such as cumulative sum control chart (CUSUM) to limit the false alarms, and threshold moving to address class imbalance in fault detection. The proposed framework is highly configurable and has been evaluated on simulated data, covering both nominal and anomalous operational scenarios. The results indicate that our solution is a promising first step, though testing with real data will be necessary to ensure that it achieves the required maturity level for operational use.
 
\end{abstract}




\end{frontmatter}

\section{Introduction}
\label{intro}
This research is part of a European effort to develop a health monitoring system (HMS) for the next generation of expandable and reusable space launchers. Currently, European launchers lack advanced onboard control or diagnostic systems, relying solely on post-flight data analysis through telemetry feeds. This limitation stems from the expendable nature of current launchers and the restricted telemetry bandwidth. However, an HMS will be essential for future reusable launchers and their engines to ensure high levels of system reliability and safety, as well as to reduce operational costs. The future HMS will need to cover both in-flight and post-flight phases, similar to existing practices in aeronautics. An overview of the usage context of the HMS in space propulsion systems is provided in \cite{ferard2021anomaly}.

As part of the onboard HMS under development, this paper introduces a data-driven approach for the fault detection and diagnosis (FDD) of the electrical valve actuators in the engine of a space launcher. This approach is designed to operate effectively during both the transient and steady-flight phases, ensuring continuous monitoring and early detection of potential issues throughout different operational stages. 

Electrical valve actuators are critical components that regulate fuel flow and other essential fluids for engine operations. These actuators typically function with precision control under high-stress conditions, given the demanding and often extreme environments involved in launch systems. They operate as part of the engine’s electrical system, which must ensure accurate and reliable performance to maintain stability and prevent potential mission failures. Given these requirements, any faults in these actuators, such as malfunctions in their electromechanical components or issues in the control signals, could impact engine performance and safety significantly.

FDD for valve actuators often relies on analyzing time series data from electrical and mechanical signals to identify any irregularities that may signal wear or failure. Within the framework of the European Space Agency's (ESA) Future Launcher Preparatory Program (FLPP), several solutions were investigated to advance the development of FDD, including for rocket engine valve actuators. While machine learning algorithms, such as neural networks, were explored, most early efforts focused on model-based approaches \citep{marcos2013fault}. More recently, artificial intelligence (AI) and machine learning (ML) methods have emerged as valuable tools in prognostics and health management (PHM) systems, bringing advanced predictive capabilities to fault detection. However, their implementation in critical systems, like those used in space launchers, remains challenging. Key barriers include ensuring data reliability, handling data limitations in fault scenarios, and maintaining interpretability and transparency in model outputs. Additionally, the integration of AI and ML solutions in complex applications demands rigorous testing and verification to ensure they can operate robustly under real-world conditions and meet stringent safety standards. Furthermore, ML algorithms like the one proposed in this paper should be designed to operate on an onboard electronic control unit (ECU).

The proposed modeling approach is implemented as a flexible framework that enables comprehensive experimentation to optimize FDD performance. This framework allows for systematic tuning of selected models, ensuring they are optimally configured to the characteristics of the provided data. In the configuration phase, we run multiple experiments to guide model selection and adjust hyperparameters effectively. The evaluated models and techniques include:

\begin{itemize}
    \item \textbf{Temporal convolutional autoencoder (TCAE)} to extract informative low-dimensional features from raw sensor data representing electrical, mechanical and valve commands signals.
    \item \textbf{Histogram-based boosting trees (HGBT)} classifiers for binary (fault detection) and multiclass (diagnosis) classification of health conditions.
    \item \textbf{Threshold moving} to address class imbalance in fault detection in fault detection by simply adjusting the threshold.
    \item \textbf{Conformal anomaly detection (CAD)} as a relatively simple and practical solution for OOD detection.
    \item \textbf{CUSUM (Cumulative Sum Control)} to reduce false positives in fault and OOD detection, enhancing reliability.
    \item \textbf{Classifier calibration} to output probabilities interpretable as confidence levels for predictions.    
\end{itemize}

To the best of our knowledge, combining TCAE with HGBT classifiers in FDD has not been explored before. Additionally, the application of CUSUM to the failure probability estimated by a binary classifier is a novel approach. Our study provides valuable insights into using techniques in the FDD field that have seen limited adoption, such as model calibration and conformal anomaly detection. In our study, we assess their benefits and limitations in enhancing the reliability of predictions and improving the overall fault detection and diagnosis capabilities.

The initial evaluation is conducted using simulated data from a physical model of an electrically controlled engine valve. This serves as a preliminary step to assess the suitability of the proposed model pipeline. In the next phase, the approach will undergo further validation using data from an integrated test bench facility with electric motors that operate real valves. In this context, the flexibility of our framework is a key advantage, allowing it to be customized to fit the specific characteristics of different datasets. 

The paper is structured as follows. Section \ref{sec:literature} reviews existing research on time series modeling and FDD, with a particular focus on electric motor applications using electrical signals such as currents and voltages. Section \ref{sec:background} outlines the key references and theoretical concepts behind the main techniques used in our framework, especially those that are less widely known. Section \ref{sec:data} presents the system under study and details the simulated data, including an analysis of class overlapping. In Section \ref{sec:method}, we describe our proposed modeling approach, highlighting the main steps of the inference process. Section \ref{sec:experiments} details the experiments conducted for model selection and hyperparameter optimization to configure the pipeline effectively. Section \ref{sec:results} evaluates the performance of the configured framework using the simulated dataset. Section \ref{sec:conclusions} summarizes the key findings, discusses limitations, and outlines potential directions for future research.

\section{Related work} \label{sec:literature}
FDD with time series data from electromechanical systems, such as electrical valve actuators, is crucial for ensuring reliability and safety in various industrial applications \citep{basak2006fault}, including in aerospace \cite{balaban2009diagnostic}. This involves analyzing sequential data patterns to identify deviations indicative of potential faults and correctly diagnosing the nature of such faults. The data's temporal characteristics are essential for capturing the dynamic behavior of the system and ensuring accurate detection and diagnosis. Electromechanical systems are equipped with sensors capable of capturing various types of signals, including current, voltage, torque, vibration, speed, and power. However, this review focus on FDD approaches that predominantly utilize electrical signals such as current and voltage.

\subsection{Classical methods} 
There are a myriad of classical data-driven methods for time series FDD developed in the fields of signal processing, statistics and machine learning. In the case of a system operating under steady-state, the fast Fourier transform (FFT) can be a reliable signal processing techniques to identify faults via spectral analysis. For non-stationary signals resulting from changing operating conditions, signal processing techniques such as short-time Fourier transform (STFT), or wavelet transform (WT) are widely used. Statistical models can be constructed by extracting statistics from the time series and finding normal boundaries \citep{choi2021deep}. Distance-based models such as Nearest Neighbors (NN) clustering method with a dynamic time wrapping distance function (DTW) are popular and often competitive techniques \cite{bagnall2017great}. More recent ensemble methods have shown impressive gains over NN-DTW, particularly COTE (Collective of Transformation-based Ensembles), which achieves strong results by combining classifiers across various time series representations \citep{bagnall2015time}. HIVE-COTE, an advanced variant, incorporates a hierarchical structure and probabilistic voting, further enhancing accuracy beyond COTE. However, this improvement comes with a significant increase in computational demands \citep{middlehurst2021hive}. Fault detection can also be achieved through predictive models by assessing the difference between the predicted and actual values, provided that the model has been trained on nominal time series data. A known predictive model is ARIMAX, an adaptation for multivariate time series of the ARIMA (Autoregressive Integrated Moving Average) model commonly used for time series forecasting. 

Signal processing and traditional machine learning techniques in fault detection for different types of electrical motor faults are reviewed in \cite{nandi2005condition}. \cite{henao2014trends} provides insights into various diagnostic techniques and discusses the integration of signal processing with machine learning for motor fault analysis. \cite{eren2004bearing} discusses signal processing methods for motor current analysis and how they are used for detecting specific faults, such as bearing damage.

Early fault detection techniques rely on analyzing the spectral characteristics of stator current. A method identifies broken rotor bar faults by examining the power spectrum of the stator current through FFT \citep{benbouzid2000review, kliman1988noninvasive}. Another well-established approach uses the negative sequence components of the stator current to detect inter-turn short circuits \citep{thomson1999line}. However, factors like voltage imbalances, specific load types, and instrument inaccuracies can induce negative sequence currents even in healthy motors, which may lead to misclassifications. The use of motor current signature analysis (MCSA) is widely used to predict rotor faults \citep{thomson2003case}, such as broken rotor bars or air gap eccentricity. By monitoring the stator current Park's vector, it is possible to detect and locate inter-turn short circuits in the stator windings of three-phase induction motors \citep{cardoso1999inter}. In \cite{AYDIN20111790}, it is introduced a two-step FDD approach for broken rotor bars faults based on sliding windows and the use of MSCA and Hilbert transforms. A recent study by \cite{allal2022diagnosis} investigates rotor faults, including broken bars and inter-turn short circuits in stator windings, through a method focused on residual harmonic analysis. In \cite{pietrzak9condition}, a methodology that combines continuous wavelet transform and classical ML algorithms such as KNN and SVM is developed. This proposed approach allows the detection and classification of inter-turn short circuits at an incipient stage. Inter-turn short circuit fault diagnosis is also studied in \cite{shih2022machine}, where two different algorithms are implemented and compared using test data from a laboratory.
The first one uses a mathematical model of the fault to assist the SVM algorithm with feature selection. This combination helps to reduce the amount of data needed for training, especially when faulty motor data are limited. It is then compared to a fault diagnosis methodology that uses convolutional neural networks to train the diagnosis model with the test data.  

\subsection{Deep-learning methods}
Classical signal processing and machine learning methods face several limitations. They rely heavily on manual feature engineering, requiring significant domain expertise, and often struggle with noise and variability in operational conditions, leading to frequent false positives or missed faults. Additionally, traditional models like SVM lack the capability to capture complex, nonlinear fault patterns and are highly sensitive to imbalances in labeled data, making it challenging to detect rare faults. Finally, these methods often lack scalability and may not generalize well to new conditions, limiting their real-world applicability. These limitations motivated the adoption of recent advancements in deep learning and hybrid approaches that combine classical techniques with deep architectures, leveraging automatic feature extraction and nonlinear modeling capabilities to better address the complexities of modern fault diagnosis. 

Deep-learning algorithms for time series classification, anomaly detection and diagnosis are reviewed in \cite{ismail2019deep, garg2021evaluation, choi2021deep}. \cite{ismail2019deep} identifies two main categories of models for time series classification: generative and discriminative. Generative models aim to learn an effective low-dimensional representation of time series before performing classification. Autoencoders (AE) are a prominent example and a widely used method in FDD \citep{qian2022review}. Discriminative models are classifiers that learn to predict labels directly from raw time series data or hand-engineered features. These features may include images or statistics generated from raw data. End-to-end deep learning models for this purpose include MLP (Multilayer Perceptrons), CNN (Convolutional Neural Networks).

A convolutional AE based approach \citep{husebo2020rapid} for the electrical faults in motor using extended Park vector. \cite{jimenez2021diagnostic} proposes a new diagnosis methodology based on discrete WT and a 1-d CNN architecture, in order to detect mechanical and electrical faults and their combination. \cite{shao2019dcnn} proposes a multi-signal fault diagnosis method that leverages the powerful feature learning ability of a CNN in images generated by WT of vibration and current signals.
Several convolutional and recurrent deep-learning architectures are evaluated in \cite{verma2020modeling, verma2023deep} to model electrical motor signals with a focus on detecting faulty measurements.

\section{Background} \label{sec:background}
In this section, we introduce specific techniques used in our modelling approach. To the best of our knowledge, TCAE and HGBT have not been applied yet for FDD in electric motors. Others techniques like calibration or conformal prediction can enhance the reliability of the predictions, but they have received little attention in the PHM community.

\subsection{Temporal convolutional autoencoder (TCAE)}
A temporal convolutional AE (TCAE) is an architecture designed for handling sequential data and originally referred to as TCN-AE by \cite{thill2021temporal}. A TCAE comprises multiple temporal convolutional layers that utilize dilated 1D-convolutional filters. In a standard 1D convolutional filter, the filter kernel slides over the input sequence at a fixed stride, processing consecutive input elements. This can limit the receptive field of the filter, making it challenging to capture long-range dependencies within the data. Dilated convolutions solve this limitation by introducing gaps between filter elements, known as dilation, which allows the filter to cover a broader range of input data points. This approach effectively expands the receptive field exponentially while maintaining the same number of parameters, enhancing the model’s ability to learn complex temporal relationships without a significant increase in computational cost.

\begin{figure}
  \centering
  \includegraphics[width=\textwidth]{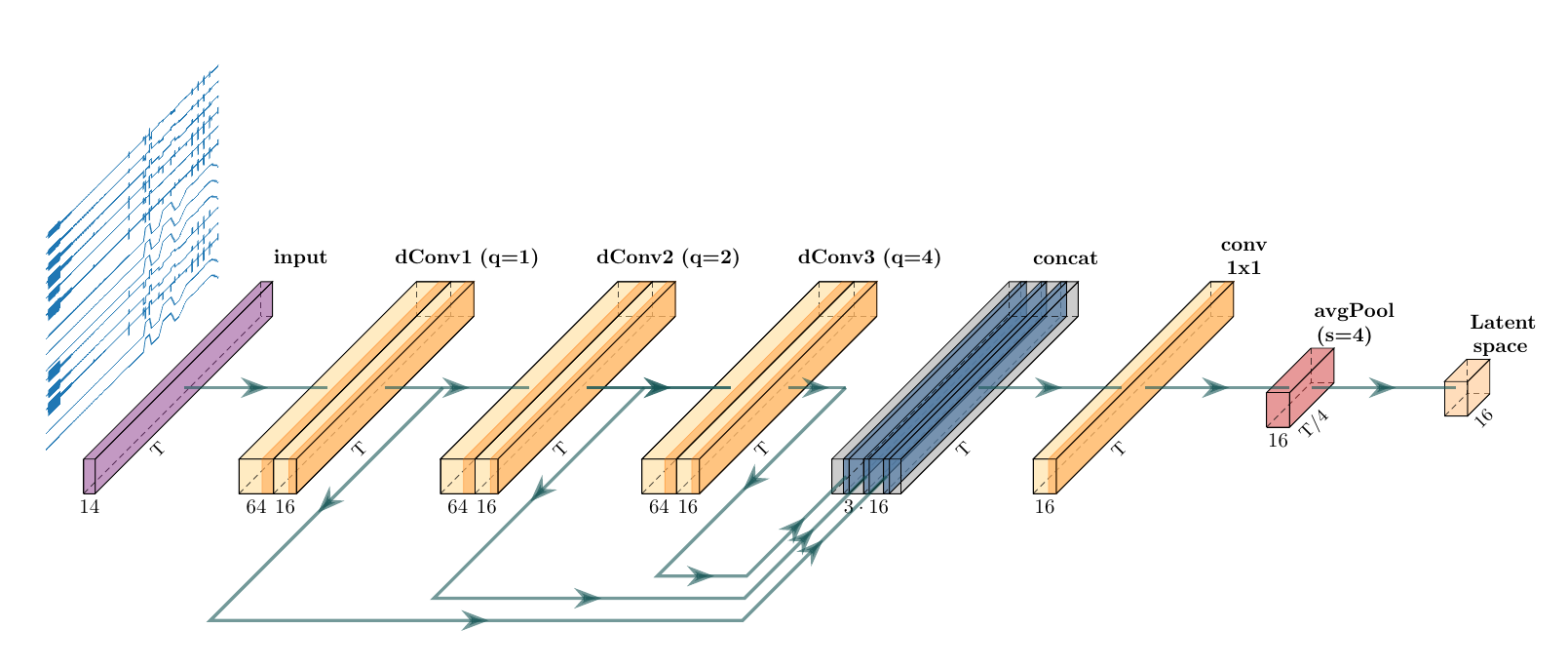}
  \caption{TCAE encoder architecture adapted from \cite{TCAE}}
  \label{fig:tcae_encoder}
\end{figure}

\begin{figure}
  \centering
  \includegraphics[width=\textwidth]{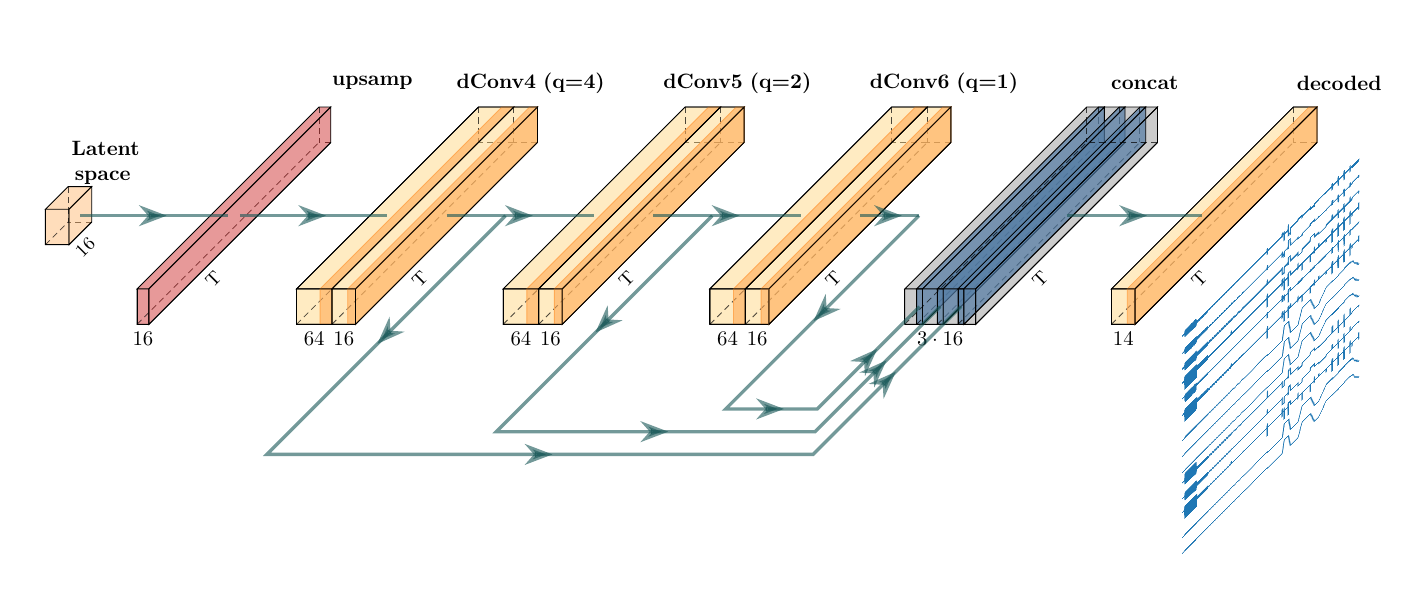}
  \caption{TCAE decoder architecture adapted from \cite{TCAE}}
  \label{fig:tcae_decoder}
\end{figure}

The final TCAE architecture employed in this paper is adapted from \citep{thill2021temporal}. It comprises an encoder and a decoder, as shown in Figures \ref{fig:tcae_encoder} and \ref{fig:tcae_decoder}. The architecture is described in greater detail hereafter.
\begin{itemize}
\setlength\itemsep{-3pt}
\item The input signal has dimensions $(c, T)$, where $c$ is the number of time series and $T$ their length. In our case, $c=14$ and $T=100$ is the size of our sliding window. In Figure \ref{fig:tcae_encoder}, the input data is depicted in purple with $c=14$. 
\item The network is then composed of $L$ dilated convolutional blocks ($L=3$ in our case), in yellow in Figure \ref{fig:tcae_encoder}. Each block is made of 64 dilated causal convolutional filters of fixed kernel size k, followed by 16 convolutional filters of kernel size 1, the activation function ReLU, and a dropout layer. The $1\times1$ convolution layers allow a fixed output size $(16, T)$, as the number of $1\times1$ filters specifies the number of channels of the output. The dilation rate has a value of 1 for the first layer and is multiplied by $b=2$, the dilation base, for each added layer.
\item The outputs of each block are then concatenated along the channel axis, allowing to identify short-term and long-term patterns. This concatenation layer using skip connections is depicted in gray in Figure \ref{fig:tcae_encoder} and its output dimension is $(16 L, T)$.
\item The resulting tensor is then compressed by reducing the number of channels to 4 with a $1\times1$ convolution and reducing the length with average pooling by a factor $s$, the sample rate. The encoded representation has dimensions $(c_{\text{latent}}, \frac{T}{s})$ ($c_{\text{latent}} = 16$ in Figure \ref{fig:tcae_encoder}).
\item In our solution, we added a convolutional layer at the end of the encoder, with input dimension $\frac{T}{s} \times c_{\text{latent}}$ and output dimension $c_{\text{latent}}$. It allows us to reduce the dimension even further without losing information.
\end{itemize}

An important architectural choice is $L$, the number of blocks. To ensure that the network covers the entire input sequence while avoiding redundant computation, we can use Equation \ref{eq:tcae} defined as:
\begin{equation}
L = \lceil log_{b}\left( \frac{(T-1)(b-1)}{2(k-1)} + 1 \right) \rceil \label{eq:tcae} 
\end{equation}
where $b=2$ is the dilation base~\citep{tcn_formula}.

The decoder is designed as a mirrored counterpart to the encoder, following the conventional structure of AE models. This mirroring includes the application of dilated convolutions with symmetric dilation rates. Specifically, the dilation rates in the decoder start at $2^{L-1}$ and decrease progressively down to 1.

\subsection{Histogram-based Gradient Boosting Trees (HGBT)}
Histogram-based Gradient Boosting Trees (HGBT) \citep{ke2017lightgbm} are a variant of the traditional gradient boosting method that is optimized for efficiency, particularly in handling large datasets. This technique is a part of the broader family of ensemble learning methods where multiple decision trees are trained sequentially, each tree aiming to correct the errors of its predecessor. In contrast to traditional boosting methods, HGBT utilizes histogram-based techniques for splitting nodes, which reduces the computational cost and speeds up the training process.

The key innovation of HGBT lies in its use of histograms to discretize continuous features into bins, allowing for faster gradient computation and reduced memory usage. This binning approach helps manage large datasets. HGBT classifiers are valued for their robustness and scalability. Their ability to handle mixed data types, manage missing values, and maintain performance with imbalanced datasets makes them a strong choice for real-world industrial and sensor-based data analysis, as confirmed in recent machine learning benchmarks.

\subsection{Classifier calibration}
Properly calibrated models produce probabilities that accurately reflect the confidence of predictions \citep{silva2023classifier}, which is particularly valuable in applications like FDD. Most classifiers produce scores $s(x) \in [0,1]$ used to rank the samples $x$ from most to least probable member of a class $c$. We have $s(x)<s(y) \Rightarrow \mathbb{P}(c|x)<\mathbb{P}(c|y)$. However, the output does not give an accurate estimate of the probability of membership to the class $s(x) \neq \mathbb{P}(c|x)$.

A classifier is said to be well-calibrated if the empirical class membership probability converges to the score as the number of samples classified goes to infinity: $\forall i \in \{1,...,n\}$, $\hat{\mathbb{P}}(c|s(x_i)=s) \xrightarrow[n\to\infty]{} s(x_i)=s$ with $n$ the number of samples classified. For instance, if we consider all samples classified with a score $s(x)=0.8$ for class $c$, 80\% of the samples should be members of class $c$. Therefore, calibrating a classifier allows us to interpret the scores as probabilities of class membership and quantify the uncertainty associated with the predictions. \cite{Zadrozny}

The calibration of a classifier can be visualized with a reliability diagram: the empirical probability $\hat{\mathbb{P}}(c|s(x)=s)$ in function of $s$. A well-calibrated classifier should yield points near the $y=x$ line. The number of possible scores being frequently too high, the score space is discretized with bins. It is important to choose the number of bins carefully.

In practice, calibrating a classifier means finding a mapping function from scores $s(x)$ to probability estimates of class membership $\hat{\mathbb{P}}(c|x)$. To avoid overfitting, a new dataset should be used for this task. It is recommended to use around 20\% of the data. The validation dataset can be used for calibration. Using the classifier output of training data to fit the calibrator would result in a biased calibrator.

Calibration methods are typically designed for binary classification tasks. To extend these methods to multiclass scenarios, the problem is often decomposed into several binary classification problems \citep{Zadrozny}. A classifier is trained for each binary problem, and calibration is applied individually. The calibrated probabilities are then combined to generate multiclass probability estimates. One common approach is one-vs-all (OVA), where a separate classifier is trained for each class, treating all other classes as negatives. The multiclass probability estimates are then derived by normalizing the calibrated probabilities of these binary classifiers.

The most widely used post-hoc calibration methods are Platt scaling and Isotonic regression. Platt scaling is a parametric method where the mapping function is a sigmoid. It was originally used to calibrate an SVM model, as the reliability curve of this model typically takes the form of a sigmoid~\citep{platt}. Isotonic regression is a non-parametric method where the mapping function is isotonic, i.e., monotonically increasing. It is the same assumption as having the samples ranked exactly by the classifier by likelihood of class membership. Isotonic regression is used to learn the mapping between scores and probability estimates~\citep{Zadrozny}.

To evaluate the calibration of the classifiers, we can use the following metrics:

\begin{itemize}
    \item The \textbf{Expected Calibration Error (ECE)} measures the calibration performance. It is defined as $\sum_{i=1}^{B} \pi_i|o_i-e_i|$ where $B$ is the number of bins, $\pi_i$ is the fraction of instances in the $i$-th bin, $o_i$ the fraction of positives samples in the bin and $e_i$ the mean calibrated probability in the bin. If the probabilities are well calibrated, the ECE will be small.
    \item The \textbf{Maximum Calibration Error (MCE)} measures the stability of the calibration method. It is defined as $\max_{i=1}^{B} |o_i-e_i|$. If a calibration method is more stable and robust, its MCE value will be smaller than other methods.
    \item The \textbf{Brier score}, also called \textbf{Mean Squared Error (MSE)}, measures the calibration performance. It is defined as $\frac{1}{N} \sum_{i=1}^{N} (y_i-p_i)^2$ where $N$ is the number of samples, $p_i$ the calibrated probability and $y_i$ the label for sample $i$.
    While a smaller MSE is preferable, it might push probability estimates towards 0 or 1~\citep{guo2017calibration}. 
\end{itemize}

\subsection{Conformal prediction}
Conformal prediction (CP) \citep{shafer2008tutorial, barber2023conformal, angelopoulos2023conformal} is a statistical technique that provides a confidence measure for classification tasks by generating prediction sets with a specified coverage probability. These prediction sets allow for uncertainty quantification by considering how well new data conforms to the model training data. CP can serve as an alternative or complement to traditional model calibration techniques. CP can enhance FDD systems by providing well-calibrated confidence scores for fault detection, improving decision-making, and reducing false alarms, especially in critical systems where reliability is essential. It enables the model to indicate when it is uncertain about a prediction, providing additional information beyond simple point predictions. In conformal anomaly detection \citep{laxhammar2010conformal}, CP can be used to assess whether a new observation is consistent with the nominal class or represents an OOD instance, making it particularly useful in identifying rare or unknown faults.

However, a limitation of conformal prediction is that its theoretical guarantees are only valid in a marginal sense, meaning that the reliability is averaged over the variability in the calibration and test samples. This marginal validity could lead to reduced accuracy in scenarios where the distribution of test data significantly deviates from that observed during calibration.

\subsection{OOD detection}
OOD detection is another important aspect that has received limited attention in FDD, despite being crucial to system reliability and safety. OOD detection allows models to identify inputs that differ significantly from the training data, which is essential for managing new or previously unseen fault types and ensuring that models do not overconfidently classify unfamiliar data. In FDD, where system anomalies or novel faults can pose significant operational risks, OOD detection provides a way to recognize potentially hazardous conditions that fall outside of known fault classes, enhancing robustness and early warning capabilities. \cite{yang2024generalized} provides an extensive overview of OOD detection, covering classic approaches and recent advancements in deep learning. 

\subsection{Post-processing with CUSUM}
Most of the FDD models discussed rely on fixed-size inputs, making sliding windows a practical solution for segmenting long sequences into manageable sections. However, window size selection is critical: overly large windows may blur or overlook short-lived anomalies, while overly small windows can limit the model ability to recognize long temporal patterns. Post-processing techniques like CUSUM \citep{page1961cumulative} may help mitigate this by tracking cumulative deviations across windows, enhancing sensitivity to shifts and allowing for more robust detection of anomalies that develop gradually over time. For instance, \cite{liu2010real} use CUSUM to accumulate small offsets to improve the detection sensitivity and raise an alarm when this sum exceeds a threshold. In case of class overlapping between nominal and failure classes, CUSUM can also be helpful to control the false alarms by differing the triggering of a fault.

\subsection{Class imbalance}
In the case of supervised learning with classifiers, class imbalance methods address scenarios where certain classes, often fault classes in FDD, are underrepresented compared to normal operating data. This imbalance can significantly hinder model performance, as classifiers tend to favor the majority class, leading to overlooked fault cases and increased false negatives. There are three main approaches to address class imbalance in classification \citep{galar2011review}. The algorithm level approaches create or modify existing algorithms to account for the significance of positive examples \citep{zadrozny2001learning}. Data level techniques such as random oversampling of the minority class (ROS), random undersampling of the majority class (RUS) \citep{zhou2005training}, synthetic oversampling of the minority class (SMOTE) \citep{chawla2002smote}. Cost-sensitive learning \citep{he2009learning} assigns higher misclassification penalties to minority class instances, effectively adjusting model sensitivity. In binary classification problems, an effective and relatively simple technique is threshold moving \citep{elkan2001foundations, zhou2005training}.

A widely used algorithm-level approach is “class weighting”, which involves adjusting a parameter during the classifier’s training process to apply a higher penalty for misclassifying the minority class. This increased penalty encourages the model to prioritize accuracy for underrepresented classes, helping to address class imbalance effectively without modifying the dataset itself.

Data-level techniques are applied as preprocessing steps to balance the dataset by adjusting the distribution of class samples. RUS reduces the majority class by randomly discarding samples until it matches the minority class, thereby minimizing dataset size and potentially losing valuable samples and increasing overfitting risks. ROS increases the minority class size by randomly duplicating its samples until it matches the majority, which may lead to overfitting. SMOTE, however, generates synthetic samples by interpolating between existing minority samples, producing a more diverse set of examples to enhance model generalization without exact duplication.

In the binary classification scenario, threshold moving is a common example of a cost-sensitive learning approach. By adjusting the default decision threshold, we can prioritize reducing false positives over false negatives, when the consequences of a false positive are deemed more critical. This strategy helps to address class imbalance by fine-tuning the decision boundaries, ensuring that the predictions better reflect the cost implications of misclassification.  

Data-level techniques often result in poorly calibrated models, typically overestimating the probability of the minority class. Class imbalance methods that rely on weighting can also introduce bias, complicating model calibration. By contrast, threshold moving effectively addresses class imbalance without negatively impacting calibration \citep{goorbergh2022, zuo2017calibration}. \cite{calib_imbalanced} shows that isotonic regression achieves more reliable calibration results than the other methods tested on imbalanced datasets. Consistent with these findings, our experiments confirm that isotonic regression provides superior calibration for binary classifiers in fault detection.

\section{System and data introduction} \label{sec:data}
\subsection{System description}
The system under consideration is a valve controlled by an electrical motor. It is made of a 3-phase electrical motor and a mechanical part to move an opening angle, e.g., between 0° and 220°. Valve movement is controlled by an electrical controller, following a set-point reference. The physical model considers a constant load torque acting on the valve. The simulation of nominal and out-of-range physical system parameters allows the modeling of different operating scenarios.

\subsection{Simulated data}
The provided simulated data for the electrical motor-controlled valve consists of trajectories, each represented by 15 distinct time series sampled at 1 kHz, as detailed in Table \ref{tab:time_series}. It is important to note that the variable \textit{Id\_Ref} remained constant and was therefore excluded from further analysis. As a result, each trajectory is represented by 14 distinct time-series variables. 

\begin{table}[h!]
\centering
\begin{tabular}{lll}
\hline
\textbf{Time series} & \textbf{Name} & \textbf{Unit} \\ \hline
D-axis reference voltage & Vd\_Ref & V \\
Q-axis reference voltage & Vq\_Ref & V \\ 
Stator pulsation & Ws & rad/s \\
D-axis reference current & Id\_Ref & A \\ 
Q-axis reference current & Iq\_Ref & A \\ 
D-axis measurement current & Id\_Meas & A \\ 
Q-axis measurement current & Iq\_Meas & A \\
Set-point mechanical position & PosM\_Set & rad \\
Reference mechanical position & PosM\_Ref & rad \\
Measured mechanical position & PosM\_Meas & rad \\
Reference mechanical speed & SpeedM\_Ref & rad/s \\
Measured mechanical speed & SpeedM\_Meas & rad/s \\
Reference mechanical torque & TqM\_Ref & N.m \\
Set-point valve position & PosV\_Set & $^{\circ}$ \\
Measured valve position & PosV\_Meas & $^{\circ}$ \\ \hline
\end{tabular}
\caption{Time-series description.}
\label{tab:time_series}
\end{table}

\begin{figure}[h]
    \centering
    \begin{subfigure}[b]{\linewidth}
        \centering
        \includegraphics[height=0.09\textheight]{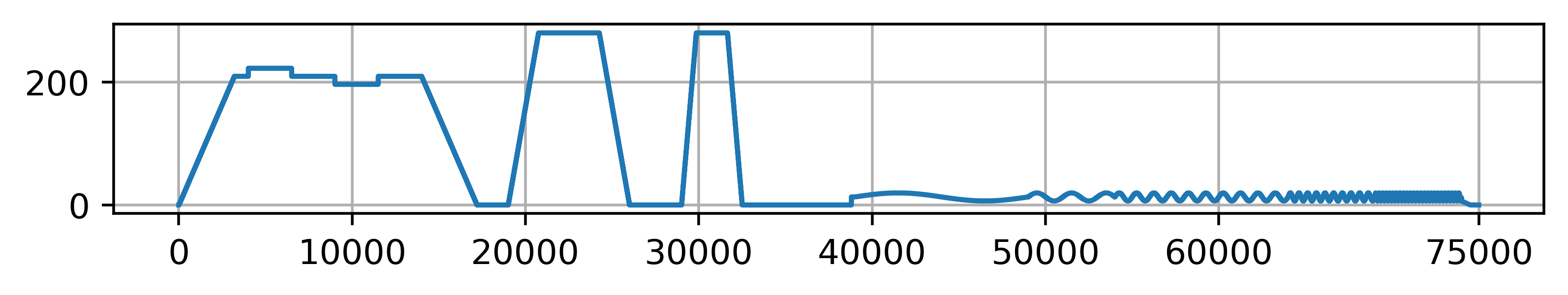}
        \vspace{-5pt}
        \caption{Type 1}
        \label{fig:traj1}
    \end{subfigure}
    \\
    \begin{subfigure}[b]{0.45\linewidth}
        \centering
        \includegraphics[height=0.09\textheight]{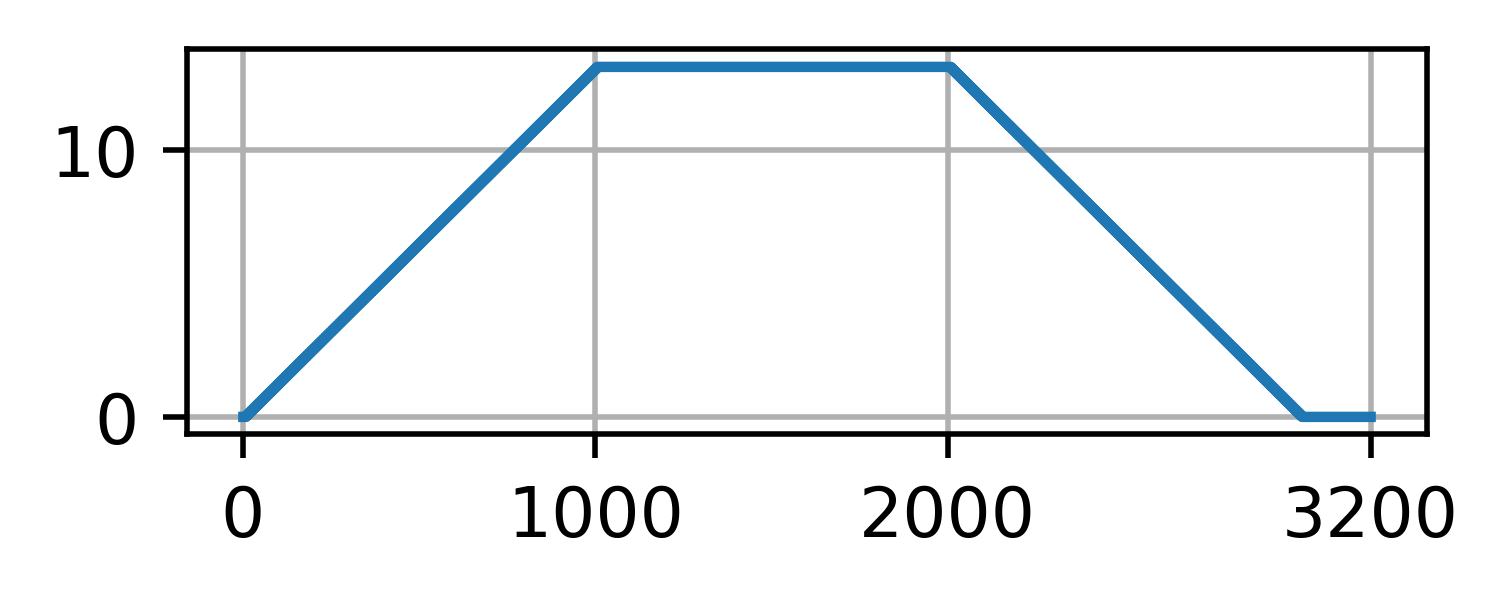}
        \vspace{-5pt}
        \caption{Type 2}
        \label{fig:traj2}
    \end{subfigure}
    \hfill
    \begin{subfigure}[b]{0.45\linewidth}
        \centering
        \includegraphics[height=0.09\textheight]{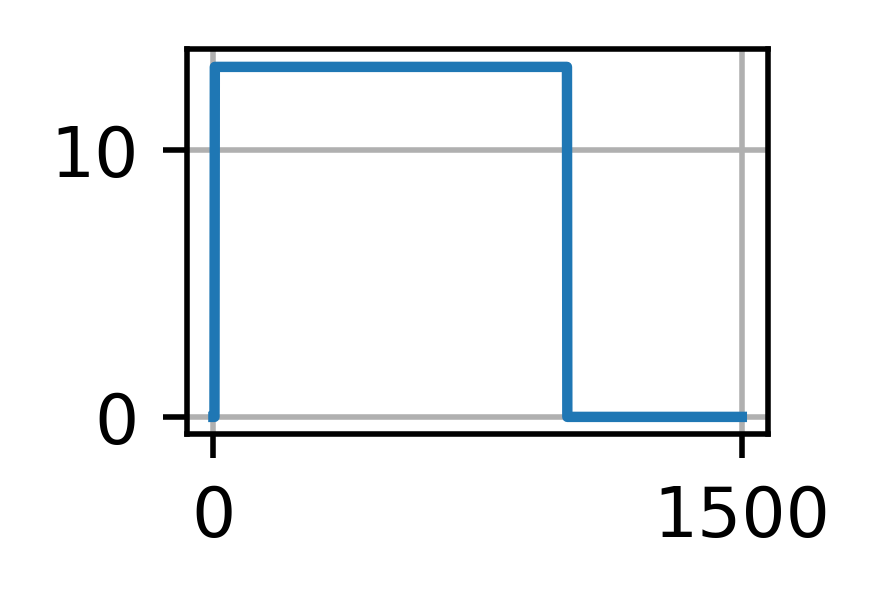}
        \vspace{-5pt}
        \caption{Type 3}
        \label{fig:traj3}
    \end{subfigure}
    \caption{Mechanical position along the trajectories of each type}
\end{figure}

Three types of trajectories are simulated, each corresponding to different valve movements. Illustrations of these trajectories can be found in Figures \ref{fig:traj1}, \ref{fig:traj2}, and \ref{fig:traj3}.

The simulation was performed using a numerical model with nine input variables. Fault types are defined based on whether one or all of these variables are outside their normal ranges. Table \ref{tab:sim_input_variables} outlines the simulation parameters and the corresponding failure labels. The failure labels are encoded in 9 bits, each bit indicating which input variable is out of range. For instance, the nominal class, where no input variables exceed their specified nominal ranges, is assigned label 0. The eleventh class, where all nine input variables are simultaneously out of range, is labeled as 511 (111111111 in binary encoding).

\begin{table}[h!]
\centering
\begin{tabular}{lllllc}
\hline
\textbf{Input variable} & \textbf{Unit} & \textbf{Nominal range} & \textbf{Anomalous range}  & \textbf{Label} \\ \hline
Main flux value & Wb & [90, 110] & [75, 85] $\cup$ [115, 125]  & 1 \\ 
Mean Inductance & H & [90, 110] & [75, 85] $\cup$ [115, 125]  & 2 \\ 
Saliency level & \% & [0, 3] & [5, 10]  & 4 \\ 
Resistance & Ohm & [100, 150] & [50, 90] $\cup$ [160, 200]  & 8 \\ 
Inertia & kg.m\textsuperscript{2} & [90, 110] & [50, 80] $\cup$ [120, 150] & 16 \\ 
Friction & N.m.s & [0, 110] & [125, 200]  & 32 \\ 
Dry friction & N.m & [0, 110] & [125, 200]  & 64 \\ 
Load Torque & N.m & [0, 110] & [125, 200]  & 128 \\ 
Unbalanced & \% & [0, 2] & [5, 20] & 256 \\ \hline
\end{tabular}
\caption{Description of the simulation input variables and their ranges}
\label{tab:sim_input_variables}
\end{table}

\subsection{Dataset preparation} \label{sec:datasets}
For experimentation, two datasets with generated simulated trajectories were provided: one designated for model development and another set aside for final validation purposes. 

In order to assess the degree of class overlapping and determine which failure cases could be distinguished, we applied various data analysis techniques to the development dataset. These included Principal Component Analysis (PCA) \citep{pearson_pca}, Uniform Manifold Approximation and Projection for Dimension Reduction (t-SNE) \citep{tsne} and Uniform Manifold Approximation and Projection for Dimension Reduction (UMAP) \citep{mcinnes2018umap}. All the analyses consistently indicated substantial overlap among classes, with only 3 out of the 10 failure cases being separable.

\begin{figure}[!h]
    \centering
    \begin{subfigure}[b]{0.49\linewidth}
        \includegraphics[width=\linewidth]{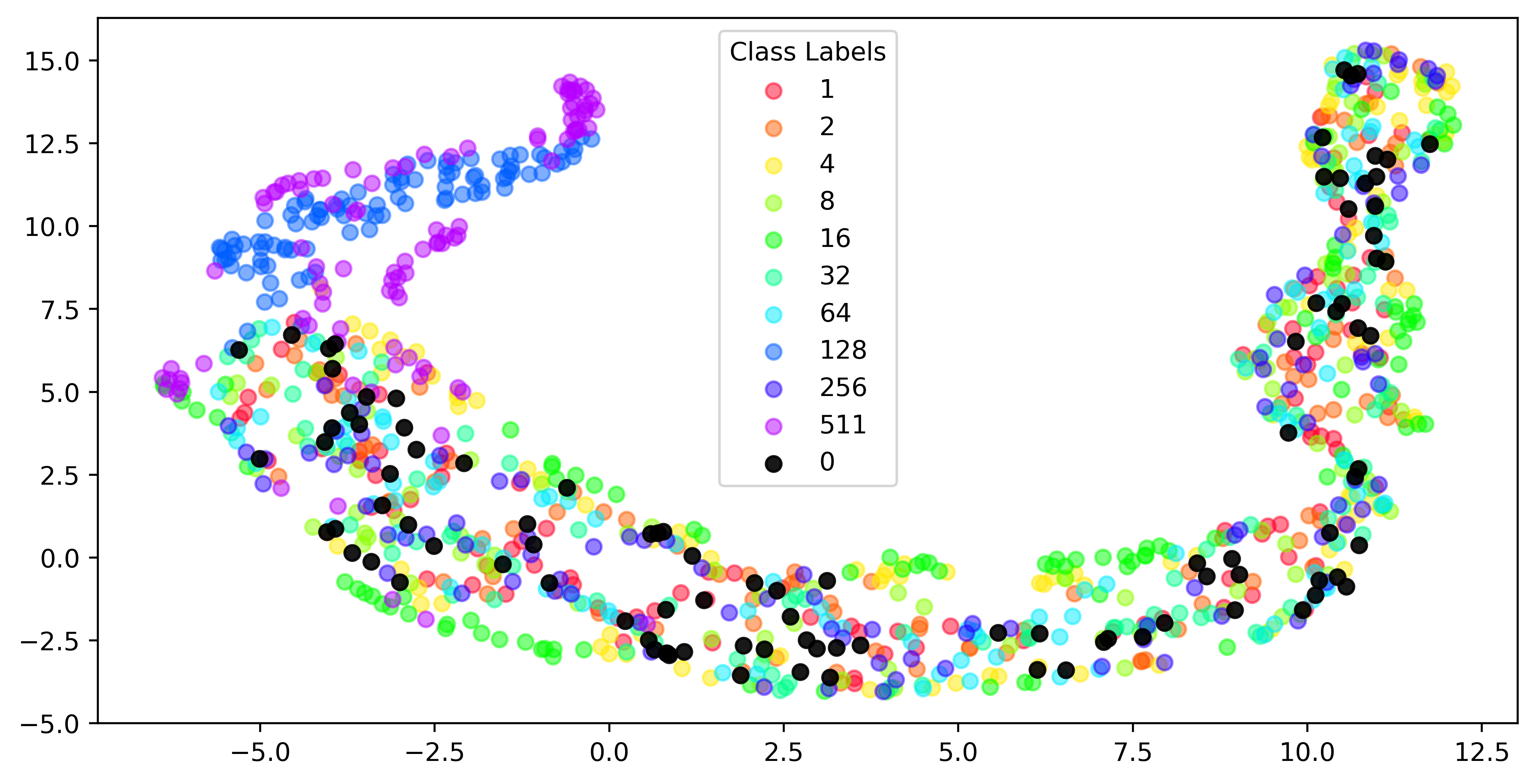}
        \caption{All classes}
        \label{fig:umap_per_class_all}
    \end{subfigure}
        \hfill
    \begin{subfigure}[b]{0.49\linewidth}
        \includegraphics[width=\linewidth]{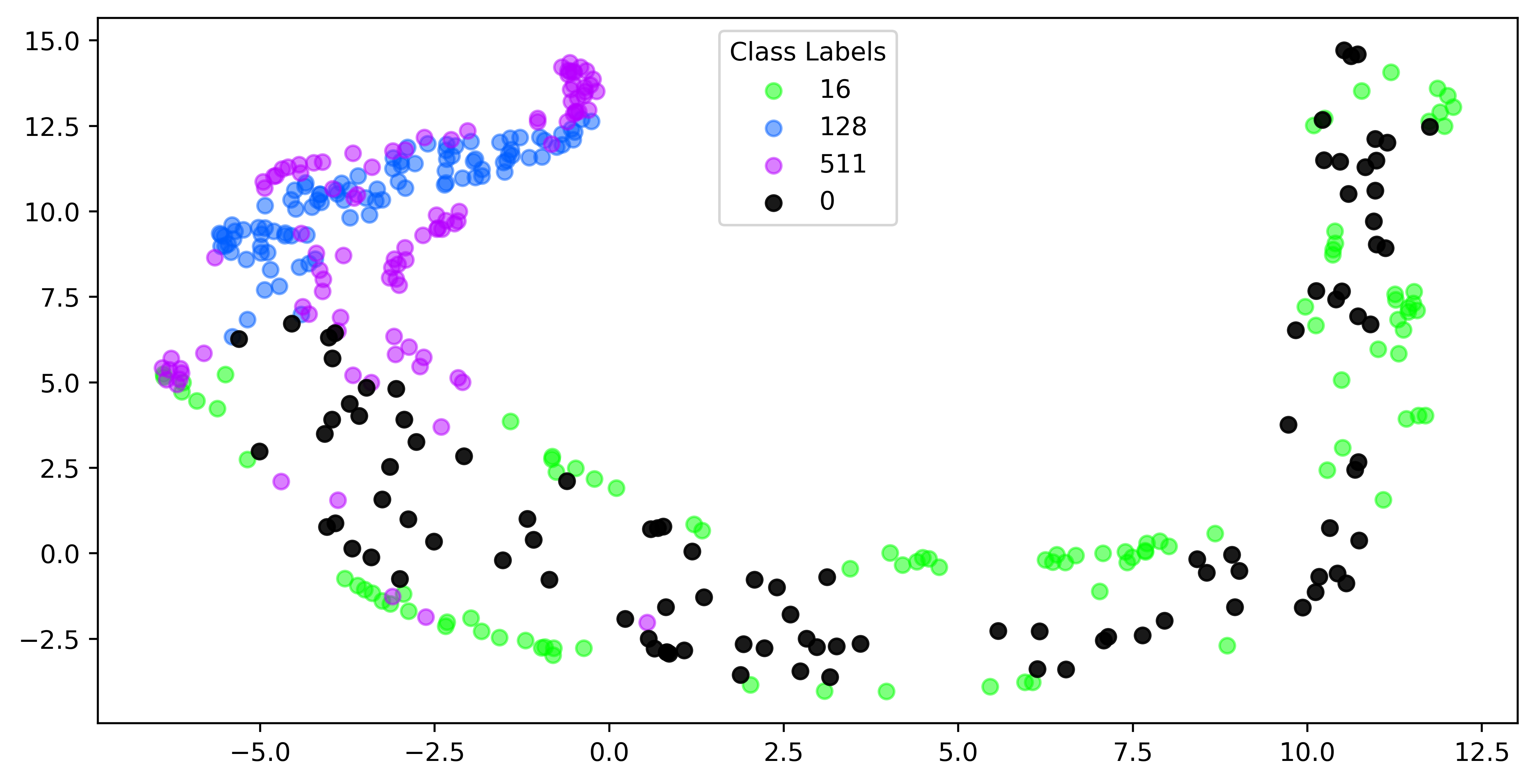}
        \caption{Subset of separable classes}
        \label{fig:umap_per_class_subset}
    \end{subfigure}
    \label{fig:umap}
    \caption{UMAP visualization of T3 trajectories to illustrate class overlapping.}
\end{figure}

Figure \ref{fig:umap_per_class_all} illustrates the issue with the UMAP method. Each point corresponds to a trajectory, for which a comprehensive set of features were extracted using the \texttt{tsfreh} \citep{tsfresh} library. These features include standard time-series statistics, along with coefficients derived from Fourier and continuous wavelet transforms. For simplicity, the visualization focuses on the T3 trajectories; however, the situation deteriorates further when all trajectories are included. Notably, the nominal trajectories overlap significantly with most of the failure cases. The few failure classes that exhibit some level of separability are shown in Figure \ref{fig:umap_per_class_subset}.

Consequently, our study will focus on the subset consisting of classes 0, 16, 128, and 511.

The development dataset was further divided into training, validation, and test sets in a 60:20:20 split ratio. The validation dataset was split into two equal subsets (50:50 ratio) to serve specific purposes. Once the trajectories were split into the different datasets, samples from each trajectory were generated using an overlapping sliding window approach. The names and uses of the various splits are defined as follows:

\begin{itemize}
    \item \textit{train}: Used for training models. Some models, such as the TCAE, are trained with specific subsets (e.g., nominal data only).
    \item \textit{val}: Employed for validation during the training process, including early stopping for models like the TCAE.
    \item \textit{val2}: Utilized during hyperparameter optimization.
    \item \textit{calibration}: Formed by combining \textit{val} and \textit{val2}, used for probability calibration.
    \item \textit{test}: A test set for independent evaluation of the models during development.
    \item \textit{test2}: The final validation dataset, reserved for testing the entire pipeline after model development is complete.
\end{itemize}

Table \ref{tab:datasets} provides further details for each set, including the number of trajectories and samples, and class prevalence. The sample count corresponds to the number of sliding windows, each configured with a size of 100 and a step of 10—parameters determined as part of the final TCAE configuration (see Section \ref{sec:final_conf_tcae}). 

\begin{table}[ht]
\centering
\begin{tabular}{lS[table-format=7.0]rrrrr}
\hline
Dataset & {\# Traj.} & {\# Samples} & 0 & 16 & 128 & 511 \\
\hline
train  & 720 & 1\,906\,320 & 0.25 & 0.25 & 0.25 & 0.25 \\
val    & 120 & 317\,720 & 0.25 & 0.25 & 0.25 & 0.25 \\
val2   & 120 & 317\,720 & 0.25 & 0.25 & 0.25 & 0.25 \\
test   & 240 & 635\,440 & 0.25 & 0.25 & 0.25 & 0.25 \\
test2  & 900 & 929\,900 & 0.85 & 0.05 & 0.05 & 0.05 \\
\hline
\end{tabular}
\caption{Number of trajectories, samples, and class ratios per dataset. Samples are sliding windows of size 100 and step 10.}
\label{tab:datasets}
\end{table}

\section{Modelling approach} \label{sec:method}

\begin{figure}
    \centering
    \includegraphics[width=0.9\linewidth]{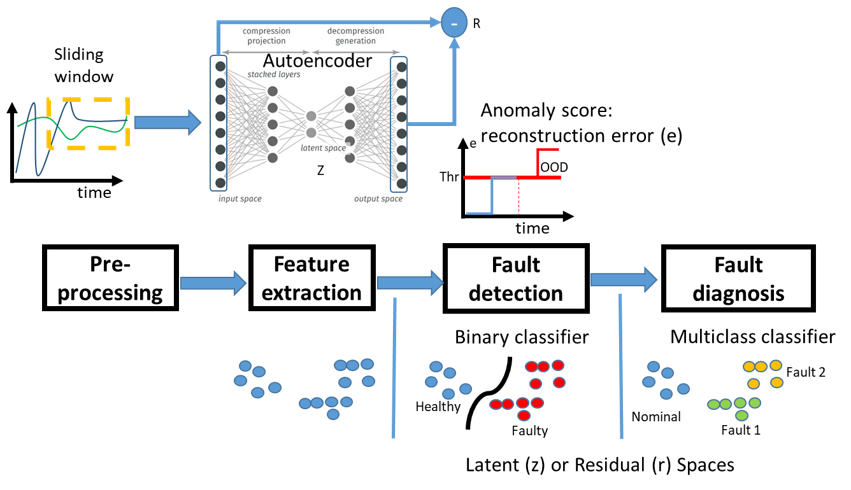}
    \caption{Method overview}
    \label{fig:method_overview}
\end{figure}

Figure~\ref{fig:method_overview}  presents an overview of the proposed approach, which integrates three key machine learning components: an autoencoder for feature extraction, a binary classifier for fault detection, and a multiclass classifier for diagnosis. The use of supervised learning is justified by the availability of labeled data corresponding to known health status classes. However, this closed-set classification assumption is unrealistic in practice. To address this, a mechanism for detecting OOD data—representing unknown fault classes or unanticipated conditions—is also integrated to enhance robustness.

In the following, we focus on the steps involved in the online phase, where the model pipeline is already configured. The configuration process is intricate and requires the design of experiments to select suitable models and fine-tune their hyperparameters. These steps are detailed in Section \ref{sec:experiments}. 

\subsection{Data preprocessing}
Data preprocessing refers to the preparation of the inputs for the subsequent task of feature extraction. It consists of data scaling and sliding windows generation.

Some machine learning models such as neural networks are sensitive to the difference in the range of the values of the input features. Therefore, we transform each feature (see Table \ref{tab:time_series}) to the \( [0, 1] \) range using min-max scaling. The scaler is fitted to the training set as part of the offline configuration process of the models.

Feature extraction is applied to samples generated from overlapping sliding windows. In the final configuration, we use a window length of 100 and a step of 10, optimizing the balance between temporal resolution and computational efficiency. Alternative window sizes have also been evaluated for their impact on performance (refer to Table \ref{tab:archi}).  

The size of the window is a critical parameter that determines the extent to which temporal dependencies within a trajectory are captured. Shorter windows often pose greater challenges for discrimination, as they further exacerbate the class overlap issues already present at the trajectory level. Conversely, larger windows enhance discriminatory power but come with trade-offs: they increase the number of parameters in the autoencoder, prolong inference time, and delay the availability of predictions. For instance, with a sampling rate of 1 kHz and a window size of 1500 points, the system must wait 1.5 seconds before generating the first prediction.

Another critical parameter is the step size, which determines the degree of overlap between two consecutive windows. A smaller step results in greater overlap, producing a higher number of samples and consequently increasing training time. Importantly, our observations suggest that the step size used during offline training must closely match the step size used in the online phase to maintain consistent performance levels.

\subsection{Feature extraction}
\begin{figure}
    \centering
    \includegraphics[width=1\linewidth]{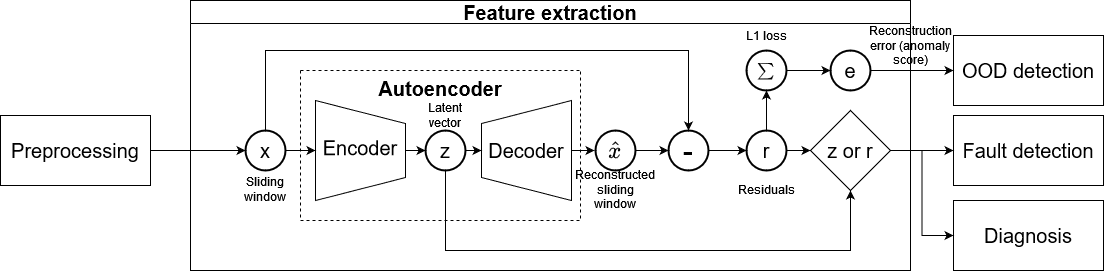}
    \caption{Feature extraction based on an autoencoder}
    \label{fig:fe_flowchart}
\end{figure}

Feature extraction is a crucial step in the entire pipeline, playing a decisive role in determining the performance of downstream tasks such as failure detection and diagnosis. As shown in Figure \ref{fig:fe_flowchart}, we employ an AE to automatically extract low-dimensional features from sliding windows of the time series data. 

Specifically, a TCAE is used (see Figures \ref{fig:tcae_encoder} and \ref{fig:tcae_decoder}), which is tailored for effective modeling and feature extraction from time series data, leveraging its structure to capture temporal dependencies efficiently. We train the TCAE exclusively on nominal data to ensure that anomalies can be detected through deviations in the reconstruction, as the network is optimized to accurately reconstruct normal data patterns. The final configuration and training parameters are detailed in Section \ref{sec:final_conf_tcae}.

The TCAE consists of two components: an encoder and a decoder. The encoder processes the input data $x$ (a sliding window of the time series) and compresses it into a lower-dimensional representation $z$, commonly known as the latent space. The decoder reconstructs the original data $\hat{x}$ from this compressed representation. The residuals are the difference between the input $x$ and the reconstructed output $\hat{x}$. By averaging the residuals across each time series in the window, we obtain $r$. The anomaly score or reconstruction error $e$ is computed as the L1 loss, specifically the mean absolute error (MAE) of the residuals. Depending on the configuration, either $r$ or $z$ can serve as inputs to classifiers for fault detection and diagnosis. On the other hand, we use $e$ for OOD detection.

\subsection{Fault detection and diagnosis}
In our approach, fault detection and diagnosis are carried out in a sequential two-step process. First, a binary classifier is employed in conjunction with the CUSUM algorithm to detect the presence of faults. If a fault is detected, a multiclass classifier is then used to diagnose the specific type of fault. This modular structure ensures that diagnosis is only performed when necessary, optimizing computational resources and reducing the risk of misclassification in nominal cases. The choice of classifiers and the calibration methods for generating confidence levels in predictions are determined experimentally during the configuration phase. (see Section \ref{sec:experiments_classifier_selection}). 

Due to having multiple simulated fault classes versus a single nominal class, the dataset shows a class imbalance in the binary classification setting. This imbalance can lead the binary classifier to favor predicting faults more often than normal conditions. To mitigate this, we compare several strategies, including algorithm-level adjustments, data-level techniques, and cost-sensitive learning approaches (see Section \ref{sec:experiments_class_imbalance}). 

\subsubsection{Inference with the binary classifier}
The low-dimensional outputs of the TCAE serve as inputs to the binary classifier (either $r$ or $z$ depending on the configuration). The classifier predicts the health status $\hat{y_{bin}}$, assigning it to class 0 (normal) or class 1 (failure), and provides a probability vector $p_{bin}$, which contains the normalized probabilities for each class. Since the binary classifier is calibrated during the offline process, the probabilities $p_{bin}$ can be interpreted as confidence levels.

\subsubsection{CUSUM and thresholding}
The binary classifier can be prone to false positives, as the prediction is based on the limited observability of a single sliding window. Thus, we apply CUSUM to accumulate the deviations to the failure probability threshold:

\begin{align}
C_i &= \max(0, \ C_{i-1} + (x_i-T_{fp}-k)) \label{eq:cusum} \\
C_0 &= 0 \nonumber
\end{align}

The variable $x_i$ is the fault probability extracted from the vector $p_{bin}$ (the calibrated probabilities output of the binary classifier), $T_{fp}$ is the failure probability threshold, $k$ is the slack variable. We trigger a fault when $C_i>T_{cs}$, where $T_{cs}$ is the CUSUM threshold. Both $T_{fp}$ and $T_{cs}$ are experimentally determined to optimize the balance between sensitivity and specificity in fault detection.

Obviously, higher values of $T_{fp}$ and $T_{cs}$ will make the fault detector more conservative and limit the false positives. Unfortunately, it will also increase the false negatives and the detection time. Therefore, there is a trade-off between these two hyperparameters which need to be considered. Please note that CUSUM is optional and can be disabled by setting $T_{cs}=0$. In that case, the algorithm will trigger a fault when $x_i>T_{fp}$.

\subsubsection{Diagnosis}
If a fault is detected, we infer the fault class and associated probability by using the off-line fitted multiclass classifier. The classifier inputs can be configured to be either the vectors r or z, the outputs of the TCAE for the current sliding window where the fault was detected.

\subsection{OOD detection} \label{sec:ood_method}
OOD data may arise from unknown faults, sensor issues (e.g., saturation or drift), or high levels of signal noise. If left undetected, such data can result in unreliable predictions, leading to potential safety hazards and operational disruptions. Our OOD detection function is designed to alert users to the increased uncertainty and risks associated with predictions on this type of data, thereby enhancing decision-making robustness. 

We implement OOD detection using an inductive conformal anomaly detection approach \citep{laxhammar2010conformal}. The OOD detector is calibrated using the reconstruction errors $e$ obtained from the calibration set. A threshold $thr_{ood}$ is defined such that a sliding window with $e>thr_{ood}$ is identified as OOD. This threshold is computed using Equation \ref{eq:thr_ood}:
\begin{equation}
thr_{ood}=\frac{\lceil(n+1)(1-\alpha)\rceil}{n}-quantile \label{eq:thr_ood} 
\end{equation} 

\noindent where $\alpha$ is the user-defined significance level controlling the type-1 error rate (false positives), and $n$ is the number of samples in the validation set.

For trajectory-level OOD detection, we observed that using a simple threshold on the number of OOD windows could lead to an excessive number of false positives. To address this, we integrated a CUSUM-based approach and introduced a user-defined threshold $thr_{ood-cs}$. This threshold specifies the minimum number of windows detected as OOD ($n_{ood}$) required to classify an entire trajectory as OOD. A warning is issued only if the trajectory satisfies the condition ($n_{ood} > thr_{ood-cs}$). This mechanism balances sensitivity and robustness, reducing false positives while promptly alerting users to potentially unreliable trajectories.

\section{Experiments} \label{sec:experiments}
The modelling approach presented in Section \ref{sec:method} is generic and needs to be configured. A global joint optimization of the whole pipeline is not possible because of the complexity due to the use of multiple models and the numerous hyperparameters. Instead, we design several experiments to optimize different parts of the process separately. The datasets used for experimentation are described in Section \ref{sec:datasets}.

\subsection{Feature extraction}

\subsubsection{TCAE architecture}
We design a set of experiments to optimize the TCAE architecture and evaluate the models with the following metrics:

\begin{itemize}
 \item the \textbf{AUROC} (Area Under the Receiver Operating Characteristic curve) measures the ability of a model to differentiate between two classes by plotting the true positive rate against the false positive rate at various thresholds. Here, we define a positive as a fault and a negative as nominal. We compute the AUROC using the reconstruction error $e$ at the window level to evaluate the TCAE ability to directly discriminate between nominal and faulty conditions without using an additional binary classifier. Additionally, we calculate AUROC for the latent space features $z$ and the residuals $r$ using a binary classifier at the window level. The average results from five HGBT models with different random seeds are reported to account for variability and ensure a robust estimate of generalization performance.
 
 \item the \textbf{accuracy} of a multiclass classification model to evaluate the fault diagnosis task at window level. It measures how often the predictions are correct. It is computed for the latent space variables $z$ and the residuals $r$. The classifier is trained on faulty data using the ground truth and not the results of the fault detection model, so we can judge the tasks independently. The mean accuracy of 5 random forest models is used, with different random seeds on the same data sets.
 
 \item the \textbf{memory footprint} of the model, which is important as we consider an onboard framework.
 
 \item the \textbf{inference time}, which is an important metric for evaluating the feasibility of real-time predictions. It is measured as the time taken to generate the decoded output of a sliding window, averaged over 100 consecutive windows. 
\end{itemize}

The set of experiments described in Table \ref{tab:archi} aims to investigate various parameters of the TCAE architecture, including different window lengths. We set the number of $k \times k$ filters to 64 as in \cite{TCAE} and the dropout rate to $0.12$.

\begin{table}
\caption{Experiments for various TCAE architectures} \label{tab:archi}
\vspace{-15pt}
\small
\begin{center}
\makebox[\textwidth][c]{%
\begin{tabular}{ccccccccccccc}
    \hline
    \rotatebox{90}{\parbox{2.4cm}{\centering \textbf{Model}}} & 
    \rotatebox{90}{\parbox{2.4cm}{\centering \textbf{Window length} $T$}} & 
    \rotatebox{90}{\parbox{2.4cm}{\centering \textbf{Encoding dimension} $c_{\text{latent}}$}} & 
    \rotatebox{90}{\parbox{2.4cm}{\centering \textbf{Number \\ of blocks} $L$}} & 
    \rotatebox{90}{\parbox{2.4cm}{\centering \textbf{Kernel size} $k$}} & 
    \rotatebox{90}{\parbox{2.4cm}{\centering \textbf{Number of \\$1\times1$ filters}}} &
    \rotatebox{90}{\parbox{2.4cm}{\centering \textbf{Sampling factor} $s$}} & 
    \rotatebox{90}{\parbox{2.4cm}{\centering \textbf{Memory footprint} (MB)}} & 
    \rotatebox{90}{\parbox{2.4cm}{\centering \textbf{Inference time} (ms)}} & 
    \rotatebox{90}{\parbox{2.4cm}{\centering \textbf{AUROC \\on $e$} (\%)}} & 
    \rotatebox{90}{\parbox{2.4cm}{\centering \textbf{Variable}}} & 
    \rotatebox{90}{\parbox{2.4cm}{\centering \textbf{AUROC binary classifier} (\%)}} & 
    \rotatebox{90}{\parbox{2.4cm}{\centering \textbf{Accuracy multiclass classifier} (\%)}} \\
    \hline
    \multirow{2}{*}{TCAE*} & \multirow{2}{*}{100} & \multirow{2}{*}{64} & \multirow{2}{*}{4} & \multirow{2}{*}{9} & \multirow{2}{*}{\textbf{32}} & \multirow{2}{*}{4} & \multirow{2}{*}{4.59} & \multirow{2}{*}{3.480} & \multirow{2}{*}{72.69} & $z$ & 85.41 & 83.03 \\
    & & & & & & & & & & $r$ & 84.57 & \textbf{92.69} \\
     \hline
    \multirow{2}{*}{TCAE} & \multirow{2}{*}{100} & \multirow{2}{*}{\textbf{64}} & \multirow{2}{*}{4} & \multirow{2}{*}{9} & \multirow{2}{*}{16} & \multirow{2}{*}{4} & \multirow{2}{*}{3.81} & \multirow{2}{*}{3.512} & \multirow{2}{*}{74.31} & $z$ & 84.24 & 83.12 \\
    & & & & & & & & & & $r$ & 84.54 & 91.73 \\
    \hline
    \multirow{2}{*}{TCAE} & \multirow{2}{*}{100} & \multirow{2}{*}{\textbf{32}} & \multirow{2}{*}{4} & \multirow{2}{*}{9} & \multirow{2}{*}{16} & \multirow{2}{*}{4} & \multirow{2}{*}{1.99} & \multirow{2}{*}{3.322} & \multirow{2}{*}{77.75} & $z$ & 85.55 & 82.51 \\
    & & & & & & & & & & $r$ & 84.36 & 91.54 \\
    \hline
    \multirow{2}{*}{TCAE} & \multirow{2}{*}{100} & \multirow{2}{*}{\textbf{16}} & \multirow{2}{*}{4} & \multirow{2}{*}{9} & \multirow{2}{*}{16} & \multirow{2}{*}{4}  & \multirow{2}{*}{1.52} & \multirow{2}{*}{3.391} & \multirow{2}{*}{76.19} & $z$ & 85.51 & 84.48 \\
    & & & & & & & & & & $r$ & 84.12 & 92.38 \\
    \hline
    \multirow{2}{*}{TCAE$^\text{(1)}$} & \multirow{2}{*}{100} & \multirow{2}{*}{16} & \multirow{2}{*}{\textbf{3}} & \multirow{2}{*}{9} & \multirow{2}{*}{16} & \multirow{2}{*}{4} & \multirow{2}{*}{1.27} & \multirow{2}{*}{\textbf{2.908}} & \multirow{2}{*}{74.77} & $z$ & 85.72 & 82.32 \\
    & & & & & & & & & & $r$ & 84.91 & 92.12 \\
    \hline
    \multirow{2}{*}{TCAE} & \multirow{2}{*}{100} & \multirow{2}{*}{16} & \multirow{2}{*}{4} & \multirow{2}{*}{\textbf{8}} & \multirow{2}{*}{16} & \multirow{2}{*}{4} & \multirow{2}{*}{1.40} & \multirow{2}{*}{3.330} & \multirow{2}{*}{76.69} & $z$ & 86.14 & 82.14 \\
    & & & & & & & & & & $r$ & 83.50 & 91.33 \\
    \hline
    \multirow{2}{*}{TCAE} & \multirow{2}{*}{100} & \multirow{2}{*}{16} & \multirow{2}{*}{4} & \multirow{2}{*}{\textbf{7}} & \multirow{2}{*}{16} & \multirow{2}{*}{4} & \multirow{2}{*}{1.27} & \multirow{2}{*}{3.392} & \multirow{2}{*}{76.05} & $z$ & 85.96 & 83.58 \\
    & & & & & & & & & & $r$ & 85.22 & 92.89 \\
    \hline
    \multirow{2}{*}{TCAE$^\text{(2)}$} & \multirow{2}{*}{100} & \multirow{2}{*}{16} & \multirow{2}{*}{4} & \multirow{2}{*}{\textbf{5}} & \multirow{2}{*}{16} & \multirow{2}{*}{4} & \multirow{2}{*}{\textbf{1.01}} & \multirow{2}{*}{3.405} & \multirow{2}{*}{69.97} & $z$ & 85.56 & 83.15 \\
    & & & & & & & & & & $r$ & 85.61 & 92.05 \\    
    \hline
    \multirow{2}{*}{TCAE} & \multirow{2}{*}{1500} & \multirow{2}{*}{\textbf{32}} & \multirow{2}{*}{7} & \multirow{2}{*}{7} & \multirow{2}{*}{16} & \multirow{2}{*}{30} & \multirow{2}{*}{2.97} & \multirow{2}{*}{7.502} & \multirow{2}{*}{77.51} & $z$ & \textbf{93.98} & 83.06 \\
    & & & & & & & & & & $r$ & 88.84 & 92.07 \\
    \hline
    \multirow{2}{*}{TCAE$^\text{(3)}$} & \multirow{2}{*}{1500} & \multirow{2}{*}{\textbf{16}} & \multirow{2}{*}{7} & \multirow{2}{*}{7} & \multirow{2}{*}{16} & \multirow{2}{*}{30} & \multirow{2}{*}{2.04} & \multirow{2}{*}{6.961} & \multirow{2}{*}{71.93} & $z$ & 92.35 & 81.11 \\
    & & & & & & & & & & $r$ & 89.55 & 90.93 \\
    \hline
    \multirow{2}{*}{TCAE} & \multirow{2}{*}{1500} & \multirow{2}{*}{16} & \multirow{2}{*}{7} & \multirow{2}{*}{7} & \multirow{2}{*}{16} & \multirow{2}{*}{\textbf{60}} & \multirow{2}{*}{1.89} & \multirow{2}{*}{7.207} & \multirow{2}{*}{\textbf{84.05}} & $z$ & 91.24 & 81.01 \\
    & & & & & & & & & & $r$ & 87.39 & 90.60 \\
    \hline
    \multirow{2}{*}{TCAE} & \multirow{2}{*}{1500} & \multirow{2}{*}{16} & \multirow{2}{*}{7} & \multirow{2}{*}{\textbf{9}} & \multirow{2}{*}{16} & \multirow{2}{*}{30} & \multirow{2}{*}{2.44} & \multirow{2}{*}{7.638} & \multirow{2}{*}{81.87} & $z$ & 91.98 & 81.63 \\
    & & & & & & & & & & $r$ & 88.47 & 91.62 \\    
    \hline
\end{tabular}
}
\end{center}
\end{table}

We experimented with 32 $1 \times 1$ filters in the TCAE* architecture, but the improvement was minimal. The architecture with 16 filters, as used in \cite{TCAE}, achieved nearly identical classification results while having a much smaller memory footprint. As a result, we set the number of $1 \times 1$ filters to 16 for the remainder of the study.

We also experimented with different encoding dimensions for the number of channels in the latent space $c_{\text{latent}}$. It appears that reducing this dimension does not significantly affect classification results. For models with windows of size 100, a reduction in encoding dimension lowers the memory footprint. However, with models for windows of size 1500, increasing the encoding dimension from 16 to 32 results in a nearly 1 MB increase in memory usage, yielding only a 1\% increase in binary AUROC and 2\% in multiclass accuracy, making this trade-off not worthwhile. Therefore, in subsequent experiments with the TCAE model, we will use an encoding dimension of 16.

On the models for windows of size 1500, we also tested a higher sampling factor which leads to a decrease of performance by 1\% on the binary classification results although the AUROC values on the reconstruction error $e$ increase by 6\%. As the results on classification are similar, we prefer the lower sampling rate to avoid losing too much information.

By comparing the fourth and fifth lines, we observe the impact of reducing the number of blocks using Equation \ref{eq:tcae}. This adjustment effectively minimizes the model's memory usage while maintaining comparable results in both binary and multiclass classification tasks. The most significant improvement is seen in the reduction of inference time.

Finally, we explored the impact of varying kernel sizes for dilated convolutions. Reducing the kernel size leads to a decrease in memory usage, without significantly affecting binary or multiclass classification performance. However, it is important to consider the trade-off between memory footprint and inference time. A smaller kernel size might necessitate additional convolutional blocks, which can increase the inference time, potentially losing the benefits of reduced memory usage.

\subsubsection{Final configuration} \label{sec:final_conf_tcae}
We have three potential TCAE candidates based on the results in Table \ref{tab:archi}. For windows of size 100, we can select the TCAE$^\text{(1)}$ architecture, which has 3 blocks and a kernel size of 9, optimized for low inference time. Alternatively, we could choose the TCAE$^\text{(2)}$ architecture with 4 blocks and a kernel size of 5, which offers a smaller memory footprint. Both architectures yield very similar classification performance. For windows of size 1500, the TCAE$^\text{(3)}$ architecture, with 7 blocks and a kernel size of 7, is a suitable option.

For the sake of simplicity and brevity, the remainder of the experiments are based solely on architecture TCAE$^\text{(1)}$. 
The specifics of this architecture are illustrated in Figures \ref{fig:tcae_encoder} and \ref{fig:tcae_decoder}.
Table \ref{tab:tcae_train} provides further details on the training parameters, including the number of epochs required for convergence and the training time for the selected TCAE$^\text{(1)}$ architecture.

\begin{table}
\centering
    \begin{tabular}{ll}
    \toprule
    Parameter & Value \\
    \midrule
    Win Step & 10 \\
    Batch Size & 4096 \\
    Loss & MSE \\
    Optimizer & Adam \\
    Learning Rate & 0.001 \\
    Early Stopping Patience & 20 \\
    GPU & NVIDIA GFORCE GTX 1080 Ti \\
    Epochs & 34 \\
    Training Time (min) & 18 \\
    \bottomrule
    \end{tabular}
\caption{TCAE$^\text{(1)}$ training configuration and convergence metrics.}
\label{tab:tcae_train}
\end{table}

\subsection{Fault detection and diagnosis}

\subsubsection{Classifier selection} \label{sec:experiments_classifier_selection}
We benchmark the following classifiers using the residuals $r$ and the TCAE$^{(1)}$ model: decision tree, SVM, logistic regression, Gaussian Naive Bayes, random forest, k-means, and HGBT. We use the \texttt{scikit-learn} library~\citep{scikit-learn} implementation with the default parameters except for the tree-based models, where the maximum depth was set to 4 to minimize training time. Each classifier is evaluated with a single run. Along with the binary AUROC and multiclass accuracy, we evaluate the calibration of the classifiers using the ECE, MCE, and MSE metrics.

\begin{table}[h!]
\centering
\makebox[\textwidth]{\begin{tabular}{ccccccccc}
\hline
\rotatebox{90}{\parbox{2.5cm}{\centering \textbf{Model}}} & 
\rotatebox{90}{\parbox{2.5cm}{\centering \textbf{AUROC binary classifier} (\%)}} & 
\rotatebox{90}{\parbox{2.5cm}{\centering \textbf{Accuracy multiclass classifier} (\%)}} & 
\rotatebox{90}{\parbox{2.5cm}{\centering \textbf{Calibration method}}} & 
\rotatebox{90}{\parbox{2.5cm}{\centering \textbf{Binary ECE}}} & 
\rotatebox{90}{\parbox{2.5cm}{\centering \textbf{Binary MCE}}} & 
\rotatebox{90}{\parbox{2.5cm}{\centering \textbf{Binary MSE}}} &
\rotatebox{90}{\parbox{2.5cm}{\centering \textbf{Multiclass ECE}}} &
\rotatebox{90}{\parbox{2.5cm}{\centering \textbf{Multiclass MCE}}} \\ \hline \hline

Gaussian & \multirow{3}{*}{74.85} & \multirow{3}{*}{59.15} & base & 0.3391 & 0.6616 & 0.3268 & 0.1840 & 0.4208 \\ \cline{4-9}
Naive & & & Platt & 0.0633 & \textbf{0.1447} & 0.1686 & 0.0578 & \textbf{0.1222} \\ \cline{4-9}
Bayes & & & isotonic & \textbf{0.0506} & 0.4213 & \textbf{0.1628} & \textbf{0.0348} & 0.1483 \\ \hline \hline

Decision & \multirow{3}{*}{80.57} & \multirow{3}{*}{66.10} & base & \textbf{0.0197} & \textbf{0.0872} & \textbf{0.1407} & \textbf{0.0320} & \textbf{0.1041} \\ \cline{4-9}
tree & & & Platt & 0.0553 & 0.2194 & 0.1505 & 0.0757 & 0.2454 \\ \cline{4-9}
& & & isotonic & 0.0519 & 0.6829 & 0.1486 & 0.0690 & 0.2148 \\ \hline \hline

Logistic & \multirow{3}{*}{83.15} & \multirow{3}{*}{64.57} & base & 0.0840 & 0.9884 & \textbf{0.1434} & \textbf{0.0334} & \textbf{0.0996} \\ \cline{4-9}
regression & & & Platt & 0.0622 & 0.3913 & 0.1516 & 0.0683 & 0.1916 \\ \cline{4-9}
& & & isotonic & \textbf{0.0494} & \textbf{0.2637} & 0.1471 & 0.0704 & 0.3218 \\ \hline \hline

Random & \multirow{3}{*}{82.53} & \multirow{3}{*}{70.27} & base & 0.0803 & \textbf{0.1104} & 0.1478 & 0.1115 & 0.3077 \\ \cline{4-9}
forest & & & Platt & 0.0656 & 0.1795 & \textbf{0.1436} & 0.0438 & \textbf{0.1241} \\ \cline{4-9}
& & & isotonic & \textbf{0.0648} & 0.1906 & 0.1446 & \textbf{0.0263} & 0.1353 \\ \hline \hline

\multirow{3}{*}{HGBT} & \multirow{3}{*}{\textbf{84.17}} & \multirow{3}{*}{\textbf{88.14}} & base & \textbf{0.0408} & 0.3590 & \textbf{0.1301} & 0.0271 & 0.1157 \\ \cline{4-9}
& & & Platt & 0.0568 & \textbf{0.3152} & 0.1382 & \textbf{0.0180} & \textbf{0.0470} \\ \cline{4-9}
& & & isotonic & 0.0574 & 0.5900 & 0.1395 & 0.0211 & 0.0724 \\ \hline

\end{tabular}}
\caption{Classifiers predictive performance and calibration}
\label{tab:perfo}
\end{table}

HGBT outperforms the other models in terms of AUROC and accuracy, as shown in Table \ref{tab:perfo}. Calibration quality varies significantly across models: for instance, the decision tree exhibits excellent calibration, while the Gaussian Naive Bayes model is poorly calibrated. The HGBT model demonstrates a reasonably good level of calibration. Furthermore, it is relatively fast to train and memory-efficient, making it our preferred choice for both fault detection and diagnosis.

\subsubsection{Class imbalance} \label{sec:experiments_class_imbalance}
We compare the effects of various class imbalance correction methods on binary classification and calibration using the residuals $r$ from the TCAE$^{(1)}$ model. An HGBT model with a maximum depth of 4 is used for binary classification. Weighting is applied using the "\textit{balanced}" option from the \texttt{scikit-learn} library, and sampling methods are implemented through the \href{https://imbalanced-learn.org}{\textit{imbalanced-learn}} library. The evaluation of classifiers is performed with a single run.

\begin{table}[h!]
\centering
\makebox[\textwidth]{\begin{tabular}{ccccccccc}
\hline
\rotatebox{90}{\parbox{2.5cm}{\centering \textbf{Method}}} &
\rotatebox{90}{\parbox{2.5cm}{\centering \textbf{AUROC} (\%)}} &
\rotatebox{90}{\parbox{2.5cm}{\centering \textbf{Precision} (\%)}} &
\rotatebox{90}{\parbox{2.5cm}{\centering \textbf{Recall} \\(\%)}} &
\rotatebox{90}{\parbox{2.5cm}{\centering \textbf{Accuracy} (\%)}} &
\rotatebox{90}{\parbox{2.5cm}{\centering \textbf{Calibration method}}} &  \rotatebox{90}{\parbox{2.5cm}{\centering \textbf{ECE}}} &
\rotatebox{90}{\parbox{2.5cm}{\centering \textbf{MCE}}} &
\rotatebox{90}{\parbox{2.5cm}{\centering \textbf{MSE}}} \\ \hline \hline

Base & \multirow{3}{*}{84.17} & \multirow{3}{*}{57.34} & \multirow{3}{*}{\textbf{90.47}} & \multirow{3}{*}{61.58} & base & \textbf{0.0408} & 0.3590 & \textbf{0.1301} \\ \cline{6-9}
model & & & & & sigmoidal & 0.0568 & \textbf{0.3152} & 0.1382 \\ \cline{6-9}
 & & & & & isotonic & 0.0574 & 0.5900 & 0.1395 \\ \hline \hline

 & \multirow{3}{*}{84.17} & \multirow{3}{*}{94.88} & \multirow{3}{*}{67.87} & \multirow{3}{*}{82.11} & base & 0.1378 & 0.2922 & 0.1561 \\ \cline{6-9}
Weighting & & & & & sigmoidal & 0.0593 & \textbf{0.2810} & \textbf{0.1388} \\ \cline{6-9}
 & & & & & isotonic & \textbf{0.0527} & 0.4673 & 0.1389 \\ \hline \hline

 & \multirow{3}{*}{\textbf{84.45}} & \multirow{3}{*}{93.63} & \multirow{3}{*}{67.82} & \multirow{3}{*}{81.60} & base & 0.1383 & 0.3226 & 0.1565 \\ \cline{6-9}
ROS & & & & & sigmoidal & \textbf{0.0568} & \textbf{0.3152} & \textbf{0.1382} \\ \cline{6-9}
 & & & & & isotonic & 0.0574 & 0.5900 & 0.1395 \\ \hline \hline

 & \multirow{3}{*}{84.24} & \multirow{3}{*}{94.30} & \multirow{3}{*}{67.78} & \multirow{3}{*}{81.84} & base & 0.1369 & \textbf{0.3073} & 0.1562 \\ \cline{6-9}
RUS & & & & & sigmoidal & \textbf{0.0568} & 0.3152 & \textbf{0.1382} \\ \cline{6-9}
 & & & & & isotonic & 0.0574 & 0.5900 & 0.1395 \\ \hline \hline

 & \multirow{3}{*}{84.22} & \multirow{3}{*}{94.86} & \multirow{3}{*}{67.94} & \multirow{3}{*}{\textbf{82.13}} & base & 0.1374 & 0.3867 & 0.1564 \\ \cline{6-9}
SMOTE & & & & & sigmoidal & \textbf{0.0568} & \textbf{0.3152} & \textbf{0.1382} \\ \cline{6-9}
 & & & & & isotonic & 0.0574 & 0.5900 & 0.1395 \\ \hline \hline

Threshold & \multirow{3}{*}{84.17} & \multirow{3}{*}{\textbf{95.53}} & \multirow{3}{*}{67.24} & \multirow{3}{*}{82.05} & base & \textbf{0.0408} & 0.3590 & \textbf{0.1301} \\ \cline{6-9}
moving & & & & & sigmoidal & 0.0568 & \textbf{0.3152} & 0.1382 \\ \cline{6-9}
 & & & & & isotonic & 0.0574 & 0.5900 & 0.1395 \\ \hline

\end{tabular}}
\caption{Model performance with various methods for class imbalance on fault detection}
\label{tab:perfo_imbalance}
\end{table}

Table \ref{tab:perfo_imbalance} shows that the threshold moving technique delivers the best classification performance without compromising model calibration.  While the other class imbalance methods yield similar classification results, they negatively affect calibration. Calibration methods generally produce the same results across all the class imbalance methods, except for class weighting. 

\begin{figure}
  \centering
  \makebox[\textwidth]{\includegraphics[width=1.15\textwidth]{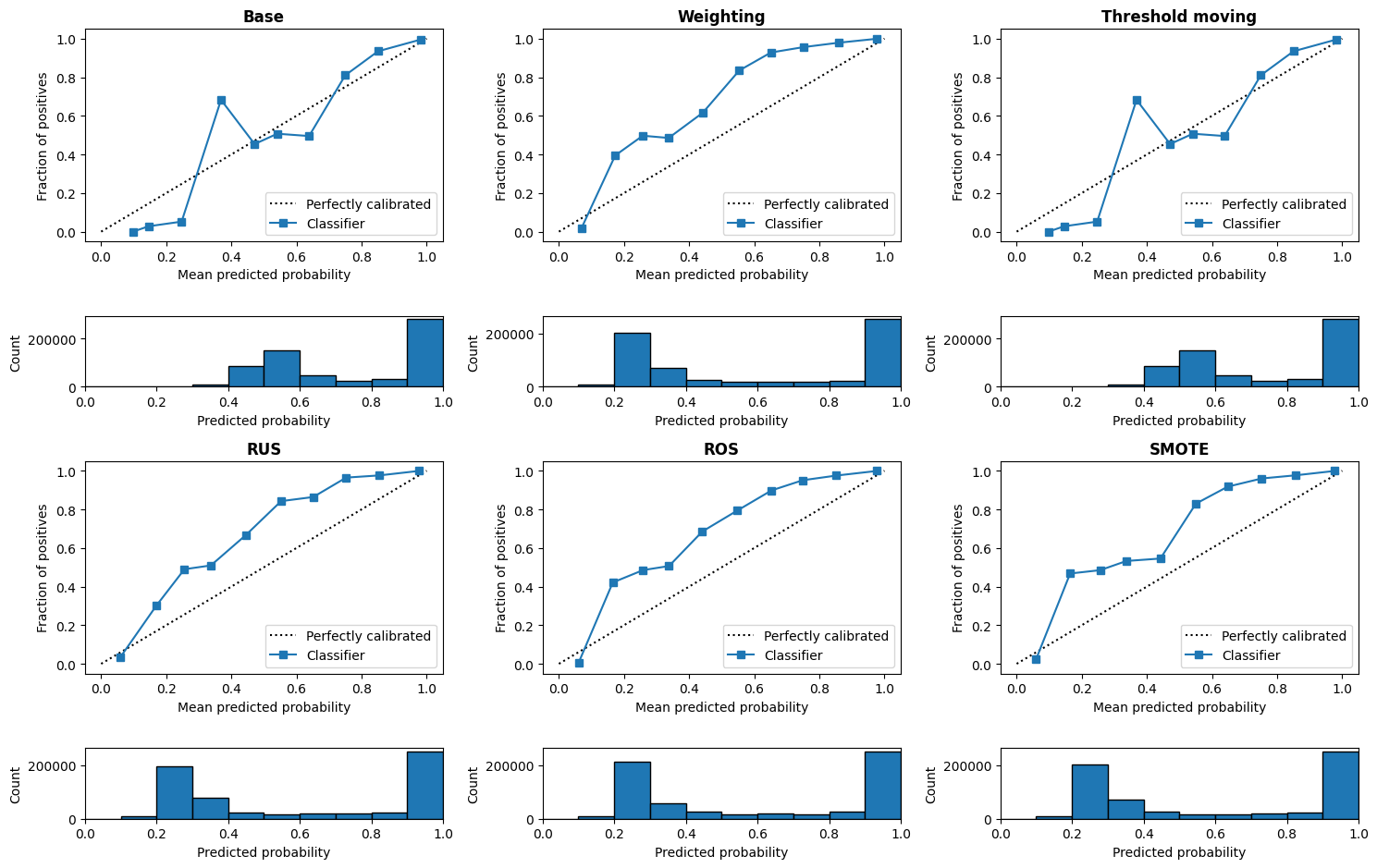}}
  \caption{Reliability plots of binary classifiers with different methods for class imbalance}
  \label{fig:rel_plots}
\end{figure}

Figure \ref{fig:rel_plots} illustrates the effect of class imbalance techniques on calibration through reliability plots. These plots reveal that weighting and sampling methods tend to produce underconfident models, with predicted scores lower on average than the actual probability of class membership. In contrast, threshold adjustment does not affect calibration; its reliability plot remains identical to that of the base model.

Therefore, it is better to use threshold moving, adjusting the probability threshold to take into account the actual prevalence of the failure (positive) cases. In our configuration with 3 failure classes, we set the threshold to 0.75 instead of the default 0.5. This shift in the probability threshold prioritizes reducing false positives over false negatives, as the impact of a false positive is considered more significant. This adjustment helps address the imbalance by fine-tuning decision boundaries, making predictions more aligned with the cost implications of misclassification.

\subsubsection{Final configuration} \label{sec:final_conf_fdd}
After selecting the HGBT classifier and the class imbalance technique, we conducted a final set of experiments to determine the configurations for the HGBT classifiers, including the input variable ($r$ or $z$). Finally, we selected the thresholds and CUSUM parameters for fault detection.

\textbf{Choice of input variable}. As presented in Table \ref{tab:archi}, the AUROC values demonstrate that both $z$ and $r$ consistently outperform the reconstruction error $e$ for fault detection. This difference is likely due to their higher dimensionality---14 for $r$ and 16 for $z$, compared to the single dimension of $e$.

Binary classification using $z$ outperforms $r$, though the difference diminishes for windows of size 100. Multiclass classification, however, consistently performs better with $r$. The latent space $z$ offers advantages in inference time and memory efficiency, since only the autoencoder’s encoder is required during deployment. In contrast, residuals $r$ provide superior interpretability, as higher values highlight greater deviations from the nominal condition. Additionally, the dimensional alignment of $r$ with time-series data allows users to identify the most anomalous physical measurements effectively.

\textbf{HGBT configuration}. Based on these results, we selected $z$ as the input for the binary HGBT and $r$ for the multiclass HGBT. The number of trees in HGBT is controlled by the \textit{max iter} parameter, which defaults to 100 in \texttt{scikit-learn}. The size of the trees can be adjusted using the \textit{max leaf nodes}, \textit{max depth}, and \textit{min samples leaf} parameters. We conducted a hyperparameter search using the \texttt{Optuna} tool, and the optimal parameters for both the binary and multiclass classifiers are presented in Table \ref{tab:hgbt_conf}. 

\begin{table}[h!]
\centering
\begin{tabular}{lll}
\toprule
Parameter & Binary HGBT & Multiclass HGBT \\
\midrule
feature & $z$ & $r$ \\
max iter & 88 & 105 \\
max depth & 6 & 9 \\
max leaf nodes & 23 & 50 \\
min samples leaf & 16 & 21 \\
learning rate & 0.05 & 0.21 \\
l2 regularization & 0.0 & 0.0 \\
\bottomrule
\end{tabular}
\caption{HGBT configurations}
\label{tab:hgbt_conf}
\end{table}

\textbf{CUSUM and thresholding}. For fault detection, configuring the CUSUM logic (Equation \eqref{eq:cusum}) is needed to better control the false positives. After conducting several experiments, we determine that setting parameters to $T_{fp}=0.75$, $T_{cs}=4$, and $k=2$ achieves an effective balance between keeping the false positive rate not higher than $5\%$ and maintaining a high recall. The threshold $T_{fp}=0.75$ reflects the prevalence of failure (positive) cases in our training set, which includes 3 failure classes. This value, higher than the default $0.5$, accounts also for the higher cost of false positives.

\subsection{OOD detection}
\subsubsection{OOD data generation}
To evaluate the OOD detection, we generated 3 new synthetic classes with 9 trajectories per class by transforming the time series of existing trajectories in the training set. For each trajectory time series $x_i$, we changed the variance and added a shift and a trend $(x_i*var+shift+i*trend)_i$. The parameters used for each class are listed in Table~\ref{tab:trajectory_data}:

\begin{table}[!ht]
\centering
\begin{tabular}{cccc}
\hline
\textbf{Class} & \textbf{Variance} & \textbf{Shift} & \textbf{Trend} \\ \hline
20             & -1                 & 5             & 0              \\ 
21             & 1                  & 5             & 0              \\ 
22             & 1                  & 5             & 1e-5        \\ \hline
\end{tabular}
\caption{Trajectory data and transformation parameters by class}
\label{tab:trajectory_data}
\end{table}

\begin{figure}[!ht]
    \centering
    \begin{subfigure}[b]{0.555\linewidth}
        \includegraphics[width=\linewidth]{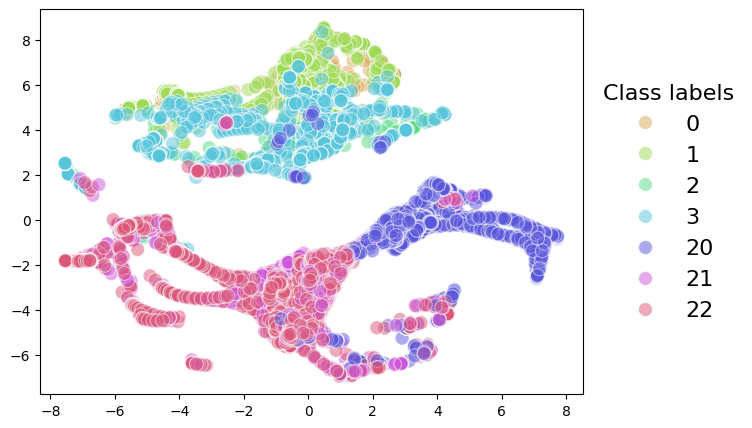}
        \caption{Latent space with OOD data}
        \label{fig:latentOOD}
    \end{subfigure}
        \hfill
    \begin{subfigure}[b]{0.435\linewidth}
        \includegraphics[width=\linewidth]{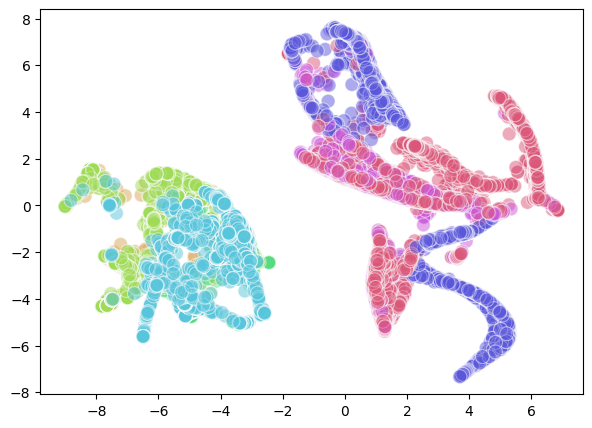}
        \caption{Residual space with OOD data}
        \label{fig:residOOD}
    \end{subfigure}
    \caption{Visualising OOD data}
\end{figure}

Figure \ref{fig:latentOOD} visualizes the synthetic OOD trajectories in the latent space $z$, and Figure \ref{fig:residOOD} in the residual space $r$. The plots reveal that the generated OOD data is clearly separated from the ID cluster in the residual space.

\subsubsection{Threshold calibration}
To determine the OOD detection threshold at the window level, the only parameter to configure is $\alpha$, the significance level. For $\alpha = 0.01$, at least 99\% of the ID are statistically guaranteed to be correctly identified. However, this guarantee assumes calibration and testing use independently generated samples from the same distribution, which is not the case here. The samples, derived from reconstruction errors, come from time-series windows, which are inherently dependent.

\begin{figure}[!ht]
    \centering
    \includegraphics[width=0.5\linewidth]{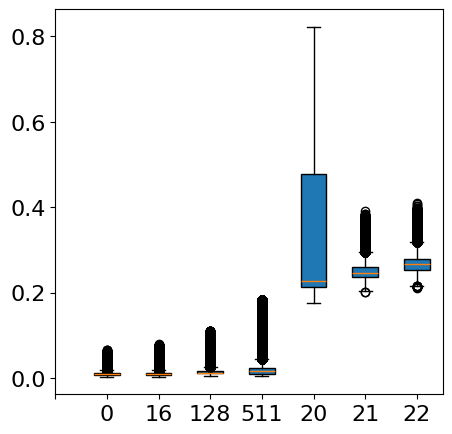}
    \caption{Boxplot of reconstruction errors per class}
    \label{fig:boxplot_error}
\end{figure}

On the calibration data, the 1\% misclassification primarily involves trajectories of types 128 and 511. This is evident in the reconstruction error $e$ distributions per class shown in Figure \ref{fig:boxplot_error}, where these two classes exhibit higher reconstruction errors. In contrast, the synthetic OOD data is consistently identified correctly, irrespective of the $\alpha$ value, due to its significantly elevated reconstruction error values.

To determine OOD data at the trajectory level, a trajectory is flagged as OOD if a specific number of its windows are identified as OOD. For $\alpha = 0.01$ and a threshold of just one window, all trajectories of classes 0, 16, and 128 are identified as ID. However, 9 out of 60 trajectories from class 511 are flagged as OOD. To ensure that all ID trajectories are correctly identified, the threshold would need to be set to at least $2\,107$ windows, which is impractically high. However, since identifying an OOD trajectory does not halt the FDD pipeline but rather alerts the user to potential unreliability in the results, a lower threshold might be more practical. This approach allows warnings to be raised more quickly, enabling proactive attention to possible issues.

\subsubsection{Final configuration} \label{sec:final_conf_ood}
We set $\alpha$ to 0.01 and calculated $thr_{ood}$ using Equation \ref{eq:thr_ood} (refer to Section \ref{sec:ood_method}) to achieve a  false positive rate at the window level not higher than $1\%$. For trajectory-level OOD detection, the CUSUM threshold ($thr_{ood-cs}$) was set to 100 sliding windows to achieve a comparable false positive rate at the trajectory level.

\section{Results} \label{sec:results}
In Section \ref{sec:experiments}, we detailed the experimental process undertaken to configure the modeling framework described \ref{sec:method}. Here, we focus on the outcomes achieved using the final framework configuration, as outlined in Sections \ref{sec:final_conf_tcae}, \ref{sec:final_conf_fdd} and \ref{sec:final_conf_ood}). The analysis covers three aspects: classification performance, detection latency, and the quality of confidence levels derived from model calibration.

The performance of the models is assessed with both the \textit{test} and \textit{test2} datasets (see Table \ref{tab:datasets}). While the \textit{test} set was used during model development, the \textit{test2} set was kept aside for final evaluation, ensuring that it was not part of the training or model selection process. The performance disparity between these two datasets serves as a measure of model generalization to unseen data. However, it is important to note that \textit{test2} exhibits a different prevalence of fault cases compared to the datasets used for model development (see Table \ref{tab:datasets}). 

\subsection{Classification performance}
Table \ref{tab:overall_perfo} presents the overall classification metrics computed at the trajectory level for the two test datasets and the three classification tasks: fault detection, diagnosis and OOD detection.

\begin{table}[ht]
\centering
\begin{tabular}{lrrrrrr}
\toprule
\textbf{Metric} & \multicolumn{2}{c}{\textbf{Fault Detection}} & \multicolumn{2}{c}{\textbf{Diagnosis}} & \multicolumn{2}{c}{\textbf{OOD Detection}} \\
\cmidrule(lr){2-3} \cmidrule(lr){4-5} \cmidrule(lr){6-7}
& \textbf{test} & \textbf{test2} & \textbf{test} & \textbf{test2} & \textbf{test} & \textbf{test2} \\
\midrule
FPR  & 0.03 & 0.06 & 0.12 & 0.08 & 0.01 & 0.00 \\
FNR  & 0.31 & 0.34 & 0.03 & 0.02 & 0.00 & 0.00 \\
Accuracy  & 0.83 & 0.80 & 0.92 & 0.94 & 0.99 & 1.00 \\
Precision  & 0.95 & 0.91 & 0.91 & 0.93 & 0.99 & 1.00 \\
Recall  & 0.69 & 0.66 & 0.93 & 0.94 & 1.00 & 1.00 \\
F1 Score  & 0.80 & 0.77 & 0.92 & 0.93 & 0.99 & 1.00 \\
\bottomrule
\end{tabular}
\caption{Classification metrics comparison between \textit{test} and \textit{test2} for FDD and OOD detection. Definitions: FPR — False Positive Rate, FNR — False Negative Rate.}
\label{tab:overall_perfo}
\end{table}

Regarding classification metrics, the overall results suggest that the models have generalized reasonably well to \textit{test2}. However, we observe a false positive rate increase, presumably due to the prevalence shift. For diagnosis and OOD detection, the results with \textit{test2} remain comparable to those obtained with \textit{test}. 

\subsubsection{Fault detection}
\begin{figure}
    \centering
    \begin{subfigure}[b]{0.49\linewidth}
        \includegraphics[width=\linewidth]{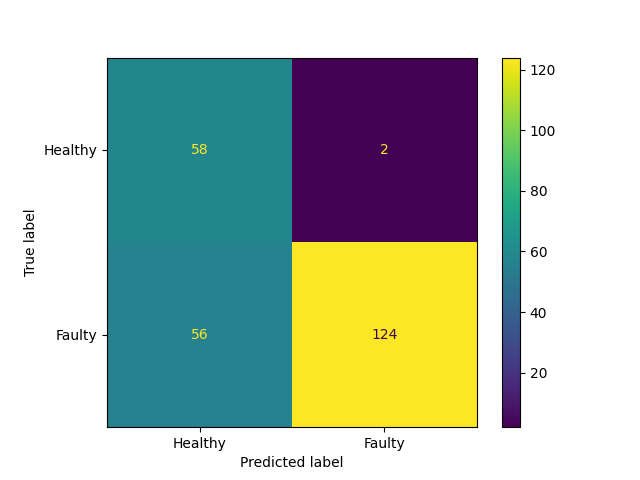}
        \caption{\textit{test}}
        \label{fig:cm_fd_test}
    \end{subfigure}
    \begin{subfigure}[b]{0.49\linewidth}
        \includegraphics[width=\linewidth]{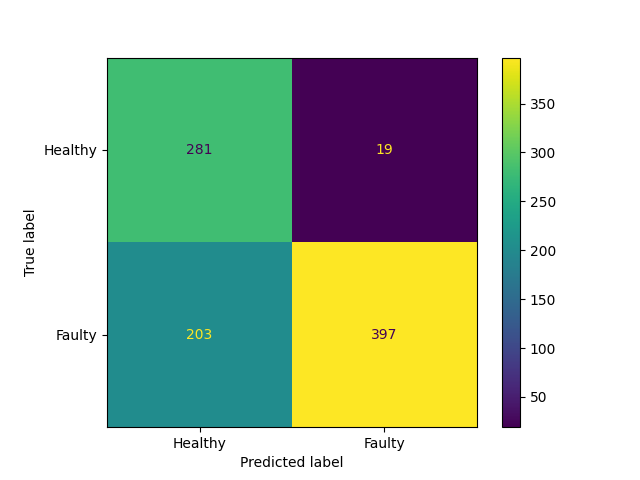}
        \caption{\textit{test2}}
        \label{fig:cm_fd_test2}
    \end{subfigure}
    \caption{Fault detection confusion matrices}
    \label{fig:cm_fd}
\end{figure}

\textbf{Overall performance}. The high precision achieved comes at the expense of a significantly lower recall, with approximately 30\% of faulty trajectories being missed (false negatives) in both the \textit{test} and \textit{test2} datasets. The confusion matrices in Figure \ref{fig:cm_fd} reveal that there are 2 false positives in \textit{test} and 19 in \textit{test2}, indicating that the false positive rate has doubled, increasing from 3\% to 6\%. 

If the prevalence shift could be estimated, it would be possible to adapt the threshold $T_{fp}$ \citep{silva2023classifier, flach2014classification} or adjust the predicted probabilities \citep{elkan2001foundations}. For example, since the model was trained with a class ratio (prevalence) of 75\% but deployed in an environment with a prevalence of 15\%, the threshold should be set to $T_{fp}=0.83$ instead of $T_{fp}=0.75$, based on the formula $r/(r + r^\prime)$ \citep{silva2023classifier}, where $r$ is the training class ratio and $r^\prime$ is the deployment class ratio. This adjustment would likely reduce the false positive rate. However, in practice, it seems difficult to estimate the deployment class ratio in our case.

\textbf{False positive analysis}. Table \ref{tab:false_positives_test_test2} presents the distribution of false positives per class and trajectory type. Most of the false positives are observed with T1 trajectories in both test sets, likely due to the increased vulnerability of CUSUM protection in these trajectories, which are characterized by their longer duration. This can be problematic during the steady flight phase, where extended periods of operation can increase the risk of false alarms.

\begin{table}
\centering
\begin{tabular}{lrrr}
\toprule
\textbf{ds} & \textbf{class} & \textbf{tt} & \textbf{count} \\
\midrule
\multirow{2}{*}{test}  & 16  & 1  & 1  \\
                       & 511 & 1  & 1  \\
\midrule
\multirow{5}{*}{test2} & 16  & 1  & 14 \\
                       & 128 & 1  & 1  \\
                       & 128 & 3  & 1  \\
                       & 511 & 1  & 2  \\
                       & 511 & 2  & 1  \\
\bottomrule
\end{tabular}
\caption{False positives in \textit{test} and \textit{test2} per predicted class and trajectory type (tt).}
\label{tab:false_positives_test_test2}
\end{table}

\textbf{Detection times}. Table \ref{tab:detection_times} shows the fault detection time statistics in seconds for \textit{test} and \textit{test2} sets. We also include the detection of the nominal class 0, which are false positives (see Table \ref{tab:false_positives_test_test2}).

\begin{table}
\centering
\begin{tabular}{llrrrr}
\toprule
class & tt & min & max & mean & std \\
\midrule
\multicolumn{6}{c}{\textit{test}} \\
\multirow{1}{*}{0}   & 1 & 0.29 & 22.20 & 11.24 & 15.49 \\
\multirow{1}{*}{16}  & 1 & 0.25 & 70.18 & 28.55 & 31.10 \\
                     & 2 & 0.55 & 0.55  & 0.55  & 0    \\
\multirow{3}{*}{128} & 1 & 0.18 & 0.20  & 0.18  & 0.01  \\
                     & 2 & 0.18 & 0.20  & 0.18  & 0.01  \\
                     & 3 & 0.18 & 0.94  & 0.25  & 0.16  \\
\multirow{3}{*}{511} & 1 & 0.18 & 0.30  & 0.20  & 0.04  \\
                     & 2 & 0.18 & 0.58  & 0.21  & 0.10  \\
                     & 3 & 0.18 & 0.33  & 0.21  & 0.05  \\
\midrule
\multicolumn{6}{c}{\textit{test2}} \\
\multirow{3}{*}{0}   & 1 & 0.21 & 29.81 & 18.22 & 12.37 \\
                     & 2 & 0.70 & 0.70  & 0.70  & 0     \\
                     & 3 & 1.20 & 1.20  & 1.20  & 0     \\
\multirow{2}{*}{16}  & 2 & 0.29 & 0.57  & 0.45  & 0.14  \\
                     & 3 & 0.96 & 1.39  & 1.17  & 0.30  \\
\multirow{2}{*}{128} & 2 & 0.18 & 0.29  & 0.18  & 0.01  \\
                     & 3 & 0.18 & 0.94  & 0.22  & 0.08  \\
\multirow{2}{*}{511} & 2 & 0.18 & 1.85  & 0.22  & 0.19  \\
                     & 3 & 0.18 & 1.18  & 0.24  & 0.13  \\
\bottomrule
\end{tabular}
\caption{Detection times for \textit{test} and \textit{test2} per class and trajectory type (tt).}
\label{tab:detection_times}
\end{table}

In the \textit{test} dataset, average detection times for classes 128 and 511 are considerably shorter than for class 16 in T1 trajectories. This disparity is due to significant class overlap between classes 0 and 16, which makes it particularly difficult for the model to differentiate between them. The false positive detections of nominal class 0 trajectories occur in average after 11s. Notably, this misclassification exclusively affects trajectories of type T1. For the \textit{test2} dataset, we observe even longer detection times for T1 trajectories. In addition, we have more false positives, including for T2 and T3 trajectories.

These longer detection times indicate that the current CUSUM configuration provides some level of protection against false positives, but it remains insufficient. However, attempts to increase CUSUM thresholds to mitigate false positives lead to a trade-off, resulting in a rise in false negatives (missed alarms).

\begin{figure}
    \centering
    \begin{subfigure}[b]{0.8\linewidth}
        \centering
        \includegraphics[width=\linewidth]{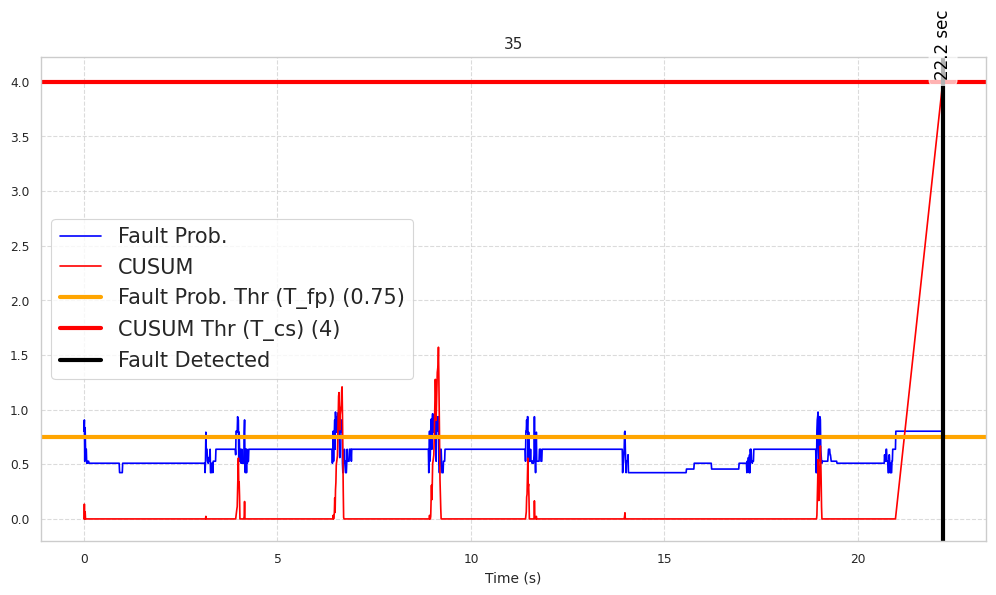}
        \caption{\textit{False positive case ($T_{fp}=0.75$})}
        \label{fig:traj_35_thr_075}
    \end{subfigure}    
    \vspace{0.5cm} 
    \begin{subfigure}[b]{0.8\linewidth}
        \centering
        \includegraphics[width=\linewidth]{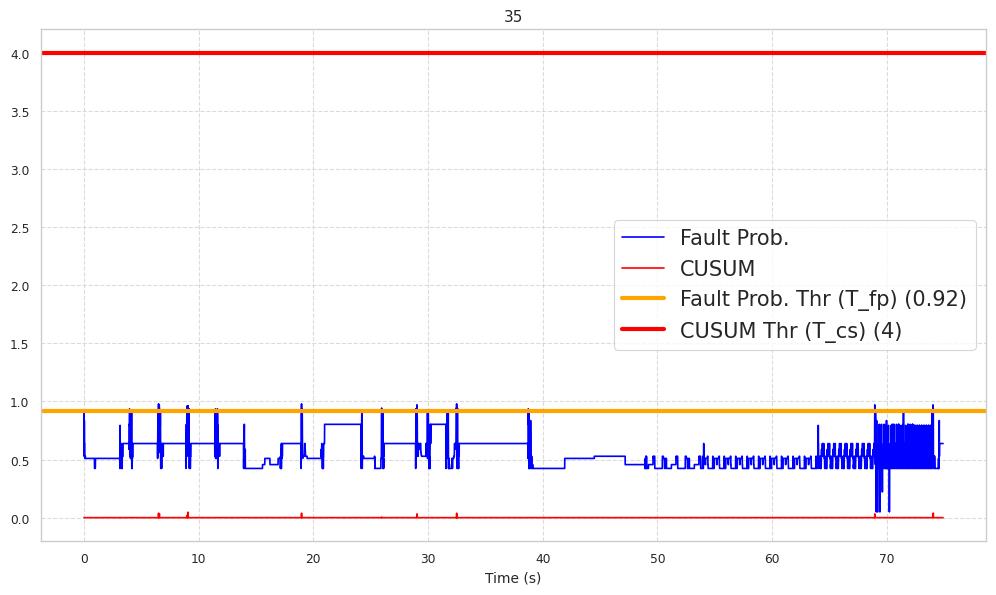}
        \caption{\textit{False positive case avoided ($T_{fp}=0.92$})}
        \label{fig:traj_35_thr_092}
    \end{subfigure}
    \caption{Example of threshold tuning to reduce FPR (Trajectory with ID=35)}
    \label{fig:thr_tuning}
\end{figure}

\textbf{Adapting the thresholds}. To further reduce the FPR, we can adjust either $T_{fp}$ or $T_{cs}$. For example, increasing $T_{fp}$ from 0.75 to 0.92 reduces the number of false positive from 21 to 1. However, this adjustment also increases FNR from approximately 30\% to 40\%. Figure \ref{fig:thr_tuning} illustrates a case where a false positive in a T1 trajectory is avoided by shifting $T_{fp}$. The plots also demonstrate how the CUSUM values (in red) evolves over time, accumulating deviations relative to $T_{fp}$. With $T_{fp}=0.75$, $T_{cs}=4$ is reached in 22.2 seconds, triggering the fault. In contrast, with $T_{fp}=0.92$, however, the deviations from $T_{fp}$ are significantly smaller, preventing CUSUM from reaching $T_{cs}=4$ and triggering the fault.

\subsubsection{Fault diagnosis}
The metrics in Table \ref{tab:overall_perfo} and the confusion matrices in Figure \ref{fig:cm_fd} indicate the multiclass classifier performs well in discriminating between the three fault classes. We have not considered class 0 in the diagnosis evaluation, because distinguishing between nominal and non-nominal cases is handled by the fault detection function. Consequently, confusion between class 0 and class 16 is not a concern for diagnosis. However, class 16 remains challenging for diagnosis, presumably due to significant overlap with the other classes.

\begin{figure}
    \centering
    \begin{subfigure}[b]{0.49\linewidth}
        \includegraphics[width=\linewidth]{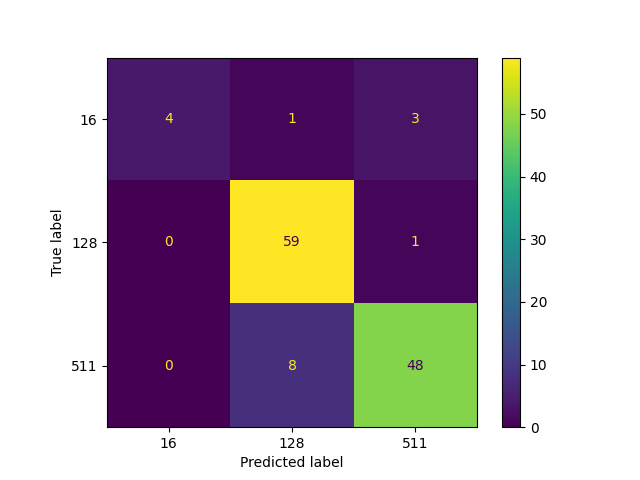}
        \caption{\textit{test}}
        \label{fig:cm_diag_test}
    \end{subfigure}
    \begin{subfigure}[b]{0.49\linewidth}
        \includegraphics[width=\linewidth]{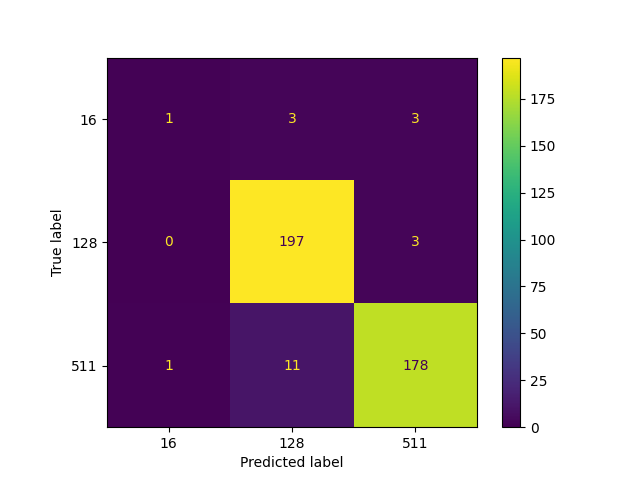}
        \caption{\textit{test2}}
        \label{fig:cm_diag_test2}
    \end{subfigure}
    \caption{Diagnosis confusion matrices}
    \label{fig:cm_diag}
\end{figure}

\subsubsection{OOD detection}
The metrics in Table \ref{tab:overall_perfo} show that the model meets the FPR target, maintaining it to a maximum of 1\%. Additionally, the model demonstrates perfect sensitivity, with a recall of 1.0, indicating that it correctly identifies all generated OOD trajectories. However, this result should be interpreted with caution. The high sensitivity could be an overestimate if the real OOD cases are more similar to the ID data than the synthetic OOD data used for testing.

\subsection{Confidence level estimation}
\subsubsection{Classifier calibration}
The quality of the confidence level is based on the proper calibration of the classifier's scores, which can be evaluated using calibration metrics and reliability plots. In the reliability plots, the bin edges are computed using the 'quantile' strategy, meaning they are based on quantiles of the predicted probabilities, ensuring an equal number of samples in each bin. For these plots, the number of bins is set to 5. Since calibration metrics can be sensitive to the number of bins used, we compute the results using four different bin sizes.

\begin{figure}
    \centering
    \begin{subfigure}[b]{0.32\linewidth} 
        \includegraphics[width=\linewidth]{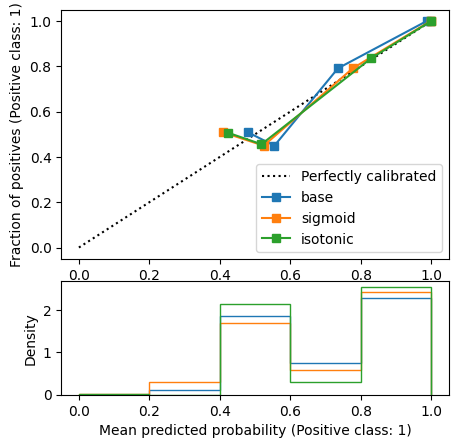}
        \caption{\textit{test}}
        \label{fig:bin_cal_test}
    \end{subfigure}
    \hfill 
    \begin{subfigure}[b]{0.32\linewidth} 
        \includegraphics[width=\linewidth]{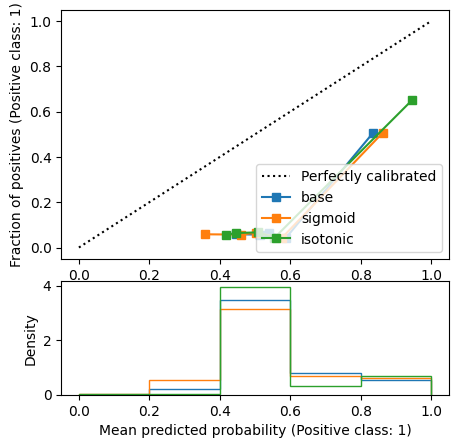}
        \caption{\textit{test2}}
        \label{fig:bin_cal_test2}
    \end{subfigure}
    \hfill 
    \begin{subfigure}[b]{0.32\linewidth} 
        \includegraphics[width=\linewidth]{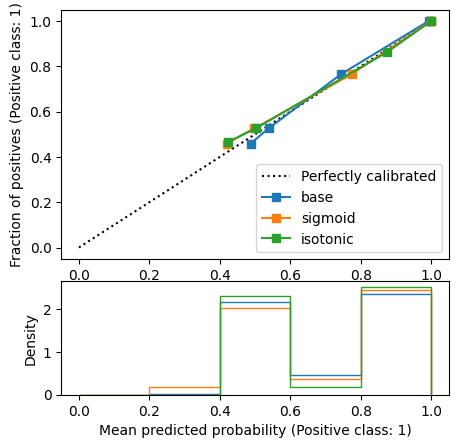}
        \caption{\textit{test2} (with \textit{test} prevalence)}
        \label{fig:bin_cal_test2_2}
    \end{subfigure}
    \caption{Calibration reliability plots for the binary classifier}
    \label{fig:bin_cal}
\end{figure}

\textbf{Binary classifier calibration}. The metrics for the binary classifier are presented in Table \ref{tab:bin_cal_metrics} and the reliability diagrams in Figure \ref{fig:bin_cal}. Although the calibration of the base model (HGBT) is fairly accurate, it can be further improved in terms of Expected Calibration Error (ECE) with isotonic calibration, though this comes at the cost of a significantly higher Maximum Calibration Error (MCE). 

\begin{table}
\centering
\begin{tabular}{rrrrrrr}
\toprule
\# Bins & \multicolumn{2}{r}{Base (HGBT)} & \multicolumn{2}{r}{Sigmoid} & \multicolumn{2}{r}{Isotonic} \\
 & ECE & MCE & ECE & MCE & ECE & MCE \\
\midrule
\multicolumn{7}{c}{\textit{test}} \\
5  & 0.041 & 0.303 & 0.021 & 0.200 & 0.008 & 0.570 \\
10 & 0.042 & 0.332 & 0.041 & 0.386 & 0.033 & 0.570 \\
15 & 0.049 & 0.329 & 0.045 & 0.511 & 0.034 & 0.570 \\
20 & 0.049 & 0.374 & 0.043 & 0.499 & 0.034 & 0.570 \\
\midrule
\multicolumn{7}{c}{\textit{test2}} \\
5  & 0.439 & 0.573 & 0.406 & 0.592 & 0.403 & 0.591 \\
10 & 0.439 & 0.583 & 0.406 & 0.646 & 0.403 & 0.666 \\
15 & 0.439 & 0.582 & 0.406 & 0.660 & 0.403 & 0.734 \\
20 & 0.439 & 0.585 & 0.406 & 0.671 & 0.403 & 0.738 \\
\bottomrule
\end{tabular}
\caption{Binary classifier calibration metrics for \textit{test} and \textit{test2}.}
\label{tab:bin_cal_metrics}
\end{table}

We observe significant miscalibration of the binary classifier probabilities in \textit{test2}, which can be attributed to the differing prevalence of failure cases between \textit{test2} and the calibration set used for model development (see Table \ref{tab:datasets}). Specifically, in the calibration set, failure cases represent 75\% of the samples, while in \textit{test2}, this ratio drops to only 14\%. When T1 trajectories are removed from \textit{test2}, the original prevalence is restored, and the miscalibration is resolved (see Figure \ref{fig:bin_cal_test2_2}). 

This issue underscores the impact of class distribution shifts on model calibration \citep{godau2023deployment}. A practical consequence of this miscalibration is the need to estimate the deployment prevalence to ensure that the predicted confidence levels are accurate in the target environment \citep{elkan2001foundations, flach2014classification, godau2023deployment}, which is challenging. 

\begin{figure}
    \centering
    \begin{subfigure}[b]{0.8\linewidth} 
        \includegraphics[width=\linewidth]{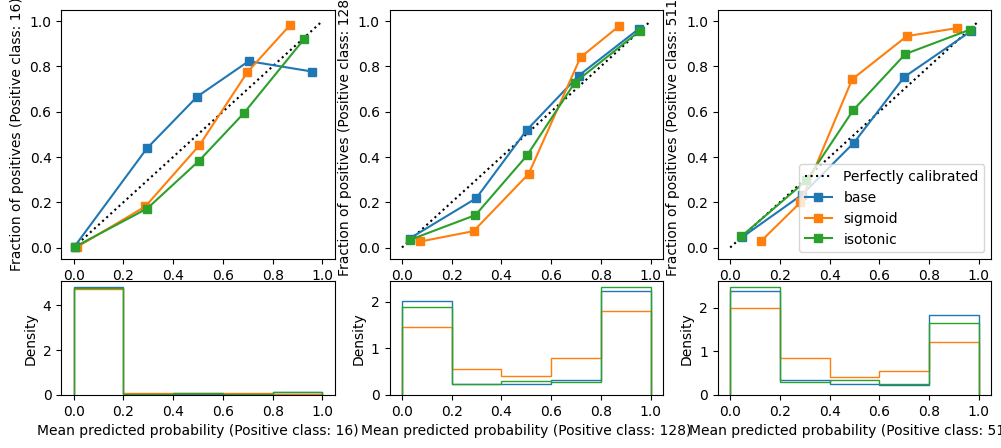}
        \caption{\textit{test}}
        \label{fig:mc_cal_test}
    \end{subfigure}
    \vspace{1em} 
    \begin{subfigure}[b]{0.8\linewidth} 
        \includegraphics[width=\linewidth]{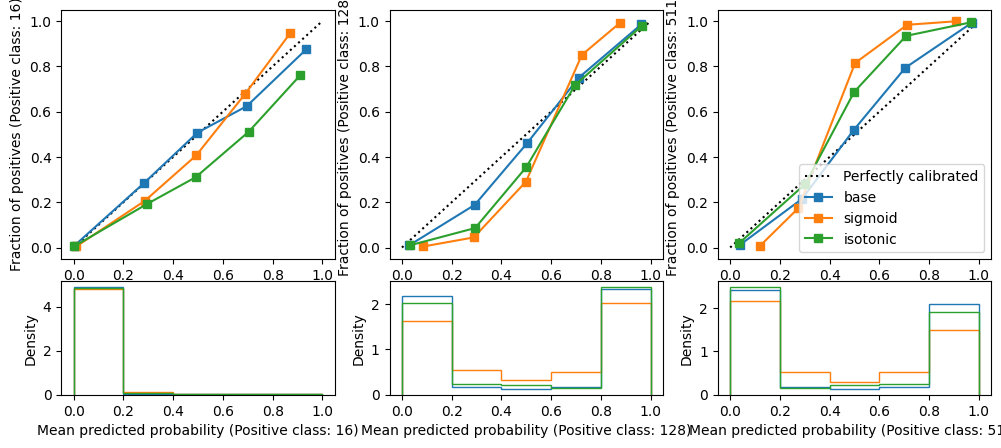}
        \caption{\textit{test2}}
        \label{fig:mc_cal_test2}
    \end{subfigure}
    \caption{Calibration reliability plots for the multiclass classifier}
    \label{fig:mc_cal}
\end{figure}

\textbf{Multiclass classifier calibration}. Figure \ref{fig:mc_cal} presents the reliability plots for each class of the base model (multiclass HGBT) and after applying the calibration algorithms. Only class 16 can be slightly improved with these algorithms, while the HGBT calibration for the other two classes is already excellent. The calibration metrics in Table \ref{tab:bin_cal_metrics} shows higher ECE and MCE values for \textit{test2}, but the resulting miscalibration is by far not as bad as it was in the binary case.

\begin{table}
\centering
\begin{tabular}{rrrrrrr}
\toprule
\# Bins & \multicolumn{2}{r}{Base (HGBT)} & \multicolumn{2}{r}{Sigmoid} & \multicolumn{2}{r}{Isotonic} \\
 & ECE & MCE & ECE & MCE & ECE & MCE \\
\midrule
\multicolumn{7}{c}{\textit{test}} \\
5 & 0.006 & 0.066 & 0.068 & 0.158 & 0.007 & 0.088 \\
10 & 0.015 & 0.066 & 0.068 & 0.170 & 0.016 & 0.094 \\
15 & 0.017 & 0.082 & 0.069 & 0.162 & 0.017 & 0.128 \\
20 & 0.015 & 0.210 & 0.072 & 0.176 & 0.020 & 0.257 \\
\midrule
\multicolumn{7}{c}{\textit{test2}} \\
5 & 0.016 & 0.084 & 0.080 & 0.196 & 0.017 & 0.125 \\
10 & 0.016 & 0.084 & 0.080 & 0.198 & 0.019 & 0.145 \\
15 & 0.016 & 0.084 & 0.080 & 0.212 & 0.019 & 0.180 \\
20 & 0.016 & 0.337 & 0.080 & 0.220 & 0.019 & 0.191 \\
\bottomrule
\end{tabular}
\caption{Multiclass classifier calibration metrics for \textit{test} and \textit{test2}.}
\label{tab:mc_cal_metrics}
\end{table}

\subsubsection{Confidence level for correct and incorrect predictions}
\begin{figure}
    \centering
    \begin{subfigure}[b]{0.8\linewidth} 
        \includegraphics[width=\linewidth]{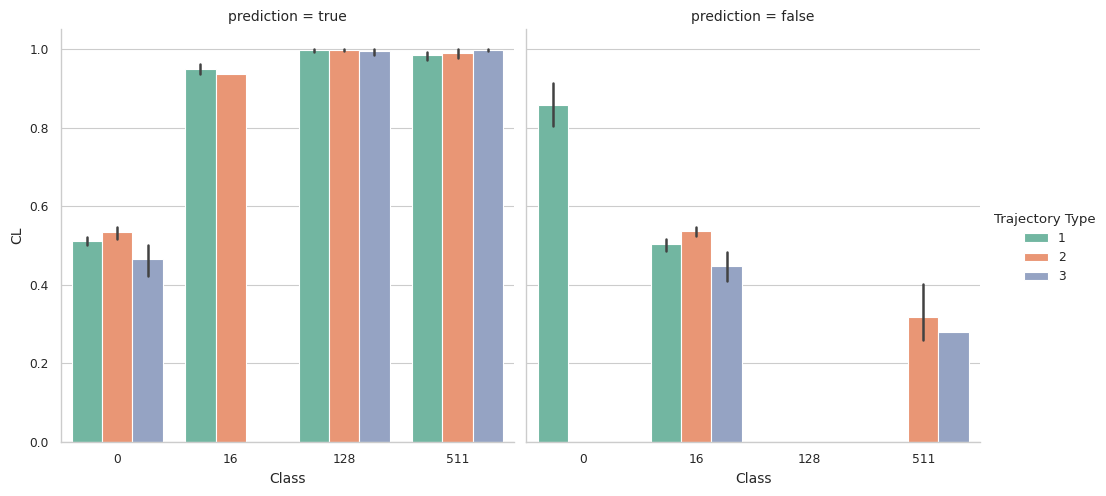}
        \caption{\textit{test}}
    \end{subfigure}
    \vspace{1em} 
    \begin{subfigure}[b]{0.8\linewidth} 
        \includegraphics[width=\linewidth]{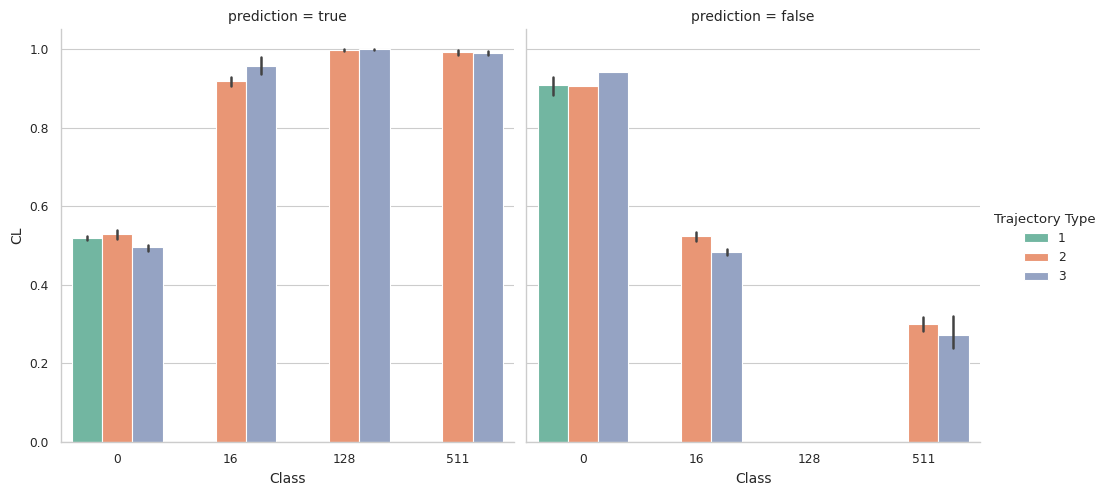}
        \caption{\textit{test2}}
    \end{subfigure}
    \caption{Confidence levels for correct (left) and incorrect (right) fault predictions}
    \label{fig:fd_cl_true_false}
\end{figure}

\textbf{Fault detection}. Figure \ref{fig:fd_cl_true_false} shows the confidence level (CL) computed as the average per fault and trajectory types over the binary calibrated probabilities. On the left plot, we show the CL for the classes correctly classified (true negatives and positives) and we expect in addition to have a high CL. The right plot focuses on the false predictions (false negatives and positives), and we want to see is a low CL indicating model uncertainty.

For both \textit{test} and \textit{test2} sets, the vast majority of trajectories in fault classes 128 and 511 are correctly detected (true positives) with a high CL. The ones missed (false negatives) have a significantly lower confidence. This means there is not enough evidence for the model to classify these trajectories as positives, but it is not sure either about them being nominal. 

The nominal class 0 is more problematic. The false positives exhibit high CL (around 0.85), which is explainable, with the current  configuration requiring the binary probability to exceed $T_{fp}=0.75$ to trigger a fault. For true negatives, CL is always lower than 0.6. This lack of confidence can be attributed to several factors. First, nominal class 0 is underrepresented in the dataset. Second, prediction uncertainty arises from overlaps with the fault classes. Lastly, it is justifiable when considering that approximately 30\% of trajectories classified as nominal are actually faulty.

Most of the overlapping seems to happen with class 16, whose true positive rate is the lowest. The false negatives in class 16 present a low CL of around 0.5 indicating model ignorance on whether these trajectories are positives or negatives.

\begin{figure}
    \centering
    \begin{subfigure}[b]{0.8\linewidth} 
        \includegraphics[width=\linewidth]{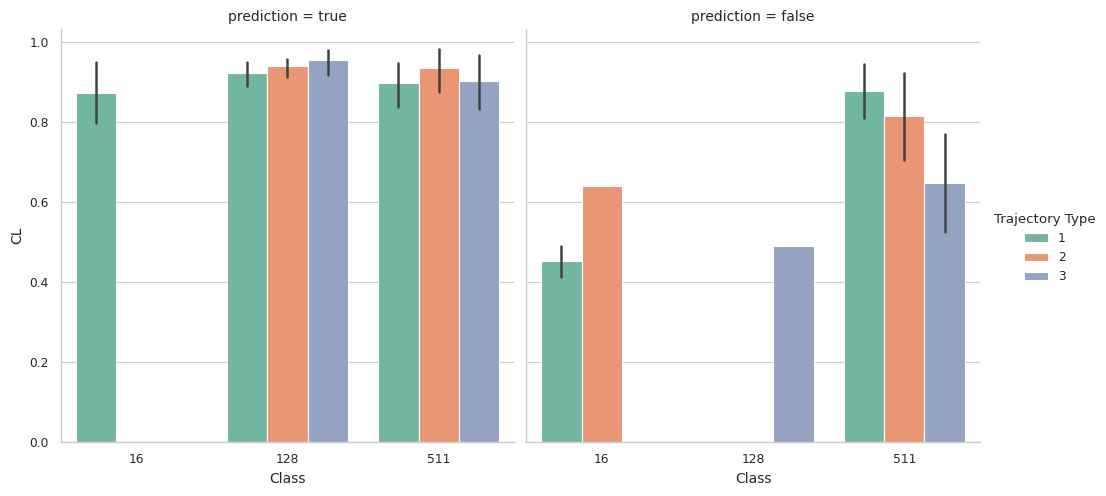}
        \caption{\textit{test}}
    \end{subfigure}
    \vspace{1em} 
    \begin{subfigure}[b]{0.8\linewidth} 
        \includegraphics[width=\linewidth]{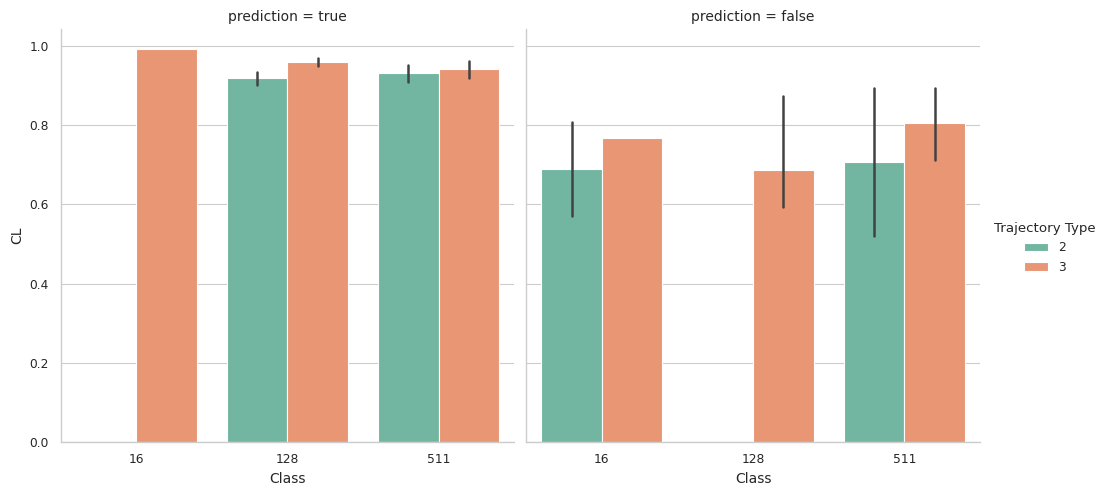}
        \caption{\textit{test2}}
    \end{subfigure}
    \caption{Confidence levels for correct (left) and incorrect (right) diagnostic predictions}
    \label{fig:diag_cl_true_false}
\end{figure}

\textbf{Diagnosis}. Figure \ref{fig:diag_cl_true_false} presents similar plots for diagnosis, where the confidence level (CL) is computed as the average of the top calibrated multiclass probabilities. It is worth noting that \textit{test2} lacks faulty T1 trajectories; consequently, these are absent from the plots. 

For the correct classifications, the average confidence level (CL) is suitably high, particularly for classes 128 and 511, where it approaches 0.9. For false predictions, the CL tends to be lower as expected, indicating some degree of model uncertainty. Additionally, there is a noticeable increase in the variance of the CL for these incorrect classifications.

\section{Conclusions} \label{sec:conclusions}
In this paper, we outline the initial steps toward developing an onboard fault detection and diagnostic solution for the engine electrical system of a space launcher. The proposed framework is highly configurable and leverages established machine learning models and statistical techniques. It integrates functions such as automatic feature extraction via TCAE, false positive control through CUSUM, fault classification with HGBT classifiers, OOD detection based on anomaly conformal prediction, and confidence level estimation for predictions with classifier calibration methods.

Model training and evaluation relies on simulated data generated from a physical model of an electrical valve. This data covers the nominal operation and ten failure scenarios, though only three fault classes are distinctly separable due to significant class overlap. An initial dataset was supplied for model development, followed by a second dataset provided at the end of development to validate the complete model pipeline. This validation dataset introduces a prevalence shift, with fault cases comprising 15\% of instances, compared to of 85\% in the development dataset.

Despite relying on established techniques, our contribution is innovative in its development of a comprehensive modeling approach for health monitoring of electrical valves. Additionally, several individual models and techniques within the framework have not been previously applied to electrical actuators. The benchmarks conducted during the framework's configuration offer valuable insights, as our work highlights common challenges in system health monitoring based on ML. 

The results obtained from the development \textit{test} set and the final validation \textit{test2} set were compared to assess model generalization. For \textit{test}, fault detection showed high precision and a low false positive rate, albeit with a high false negative rate. However, performance declined with \textit{test2}, most notably with a false positive rate that doubled. In contrast, both diagnosis and OOD detection demonstrated consistent performance across both datasets, suggesting the change in the class ratios had a minor impact on these tasks. 

We found that the shift in class prevalence led to a performance drop in fault detection and a significant miscalibration in the binary classifier. The issue of class distribution changes is discussed in the literature, with several approaches proposed to address it \citep{elkan2001foundations, flach2014classification, godau2023deployment}. The proposed approaches to mitigate this effect is to recalibrate the models or adjust the predicted probabilities or decision thresholds to account for the new prevalence. However, this adjustment is only feasible when the new prevalence can be determined, which seems challenging in our case. 

To estimate confidence levels in fault detection and diagnosis (FDD), the application of conformal prediction should be investigated. Its ability to provide uncertainty quantification with statistical guarantees under minimal distributional assumptions may make it more resilient to shifts in data distribution compared to traditional calibration techniques. 

Our contribution underscores the challenges of achieving robust model training and evaluation, which are often constrained by the reliance on simulated data in the absence of real data. To address these limitations, the next phase of this research will transition to training and evaluating the framework with real data. This data will be gathered from a purpose-built test bench equipped with cost-effective electric motors, designed to emulate the valve actuators commonly used in space launcher engines. This transition is intended to validate the framework under more realistic operating conditions. Additionally, the potential for leveraging both simulated and real data will be explored.

\section{Acknowledgements}
This document was produced as part of the ENLIGHTEN program, funded by Horizon Europe under Grant Agreement No. 101082326. The authors gratefully acknowledge the support of ArianeGroup for initiating the activity and providing the data used in the illustrations.


\bibliographystyle{elsarticle-harv} 
\bibliography{biblio.bib}
\end{document}